\pdfoutput=1
\documentclass[11pt]{article}

\usepackage[preprint]{acl}
\usepackage{subcaption}
\usepackage{times}
\usepackage{latexsym}
\usepackage{booktabs}
\usepackage{amsmath}
\usepackage{algorithm}
\usepackage{algorithmic}
\usepackage{pifont}
\usepackage[T1]{fontenc}
\usepackage[utf8]{inputenc}

\usepackage{microtype}

\usepackage{inconsolata}

\usepackage{graphicx}

\title{Can Large Vision-Language Models Detect Images Copyright Infringement from GenAI?}

\author{
\vspace{2mm} 
        Qipan Xu$^1$\hspace{3mm}
        Zhenting Wang$^1$ \hspace{3mm}
        Xiaoxiao He$^1$ \hspace{3mm}
    \\
    \textbf{
        Ligong Han$^1$ \hspace{3mm}
        Ruixiang Tang$^1$ \hspace{3mm}
    }
    \vspace{2mm}
    \\
    \hspace{-3mm}
    \textsuperscript{1}Rutgers University
    \\
    \hspace{-8mm}
}

\begin{document}
\maketitle
\begin{abstract}
Generative AI models, renowned for their ability to synthesize high-quality content, have sparked growing concerns over the improper generation of copyright-protected material. While recent studies have proposed various approaches to address copyright issues, the capability of large vision-language models (LVLMs) to detect copyright infringements remains largely unexplored. In this work, we focus on evaluating the copyright detection abilities of state-of-the-art LVLMs using a various set of image samples. Recognizing the absence of a comprehensive dataset that includes both IP-infringement samples and ambiguous non-infringement negative samples, we construct a benchmark dataset comprising positive samples that violate the copyright protection of well-known IP figures, as well as negative samples that resemble these figures but do not raise copyright concerns. This dataset is created using advanced prompt engineering techniques. We then evaluate leading LVLMs using our benchmark dataset. Our experimental results reveal that LVLMs are prone to overfitting, leading to the misclassification of some negative samples as IP-infringement cases. In the final section, we analyze these failure cases and propose potential solutions to mitigate the overfitting problem.
\end{abstract}

\section{Introduction}
The rapid advancement of generative artificial intelligence (GenAI) has ushered in a new era of content creation, enabling the synthesis of high-quality text, images, and multimedia content at an unprecedented scale. While these innovations have expanded creative possibilities and applications across industries, they have also raised significant ethical and legal concerns, particularly regarding intellectual property (IP) rights \cite{sag2023copyright, polandgenerative}. One of the most pressing issues is the unauthorized reproduction of copyrighted material, where generative models may inadvertently produce outputs that closely resemble or replicate IP-protected content \cite{zirpoli2023generative, dzuong2024uncertain, sag2023copyright, polandgenerative,wang2024evaluating}. This issue has led to widespread debates among legal experts, policymakers, and AI researchers on the potential liabilities and regulatory measures required to address copyright infringement in AI-generated content.

Existing efforts to mitigate copyright concerns in generative models have primarily focused on two key approaches: \ding{172} reducing memorization during training using techniques such as differential privacy \cite{dwork2014algorithmic}, which limits the retention of specific data points to prevent models from reproducing protected content \cite{abadi2016deep, chen2022dpgen, dockhorn2022differentially}, and \ding{173} employing prompt engineering strategies that discourage the generation of IP-infringing material through explicit negative prompts \cite{wang2024evaluating, he2024fantastic} or optimized safe prompt modifications \cite{chin2023prompting4debugging, rando2022red}. While these approaches offer some level of control over generative outputs, they do not directly address the challenge of detecting copyright infringement in already-generated content. As a result, there is an urgent need for robust evaluation methods and benchmarks to assess the ability of AI models—specifically large vision-language models (LVLMs)—to identify potential instances of copyright violations.

Vision-language models (VLMs), which integrate both textual and visual data to enable cross-modal reasoning, have demonstrated remarkable capabilities in tasks such as image classification, visual question answering (VQA) \cite{antol2015vqa}, and multimodal understanding. Notable VLMs such as CLIP \citep{pmlr-v139-radford21a}, large vision-language models (LVLMs) such as GPT-4o \cite{gpt4v}, Claude 3.5 \cite{Claude3.5}, VILA-2.7b \citep{lin2023vila}, and Qwen-VL \cite{bai2023qwen} have been trained to interpret and generate content based on textual and visual inputs, making them prime candidates for assessing IP infringement detection. However, despite their extensive deployment in various applications, the effectiveness of LVLMs in identifying copyright-protected content remains largely untested. Given the increasing reliance on these models in content moderation, digital rights management, and automated compliance monitoring, it is crucial to evaluate their ability to detect copyright infringement.

To address this gap, our work presents a systematic evaluation of LVLMs for copyright detection by constructing a dedicated benchmark dataset. Recognizing the absence of a comprehensive dataset that includes both clear cases of IP infringement and ambiguous non-infringing samples, we create a dataset comprising:
\begin{itemize}
    \item Positive samples that contain well-known IP characters generated using different AI models with direct and descriptive prompts that replicate their distinctive features.
    \item Negative samples that resemble IP characters in certain aspects but do not fully qualify as copyright violations. These images are generated using modified negative prompt engineering techniques. 
\end{itemize}
These images are selected through rigorous human annotation after the generation.

Our dataset focuses on five widely recognized fictional characters: Iron Man, Batman, Spider-Man, Superman, and Super Mario, ensuring a balanced representation of both positive and negative samples. To evaluate the effectiveness of VLMs, we conduct experiments using in-context learning (ICL)  \cite{mann2020language, dong2022survey} and zero-shot learning (ZSL) \cite{wang2019survey} approaches, where models are tested on their ability to classify image samples accurately.

Our findings indicate that while LVLMs exhibit strong recall in detecting potential copyright violations, they often suffer from overfitting and exhibit a tendency to classify ambiguous samples as infringing content, leading to a high rate of false positives. This issue highlights a fundamental challenge in using LVLMs for automated copyright detection—these models may prioritize superficial visual similarities rather than deeper conceptual understanding of IP infringement. To address this limitation, we propose a set of mitigation strategies, including contrastive learning techniques that refine the models’ ability to differentiate between genuine IP violations and non-infringing variations.

The contributions of our work are fourfold:

\begin{itemize}
    \item Introduction of a novel benchmark dataset specifically designed to evaluate the copyright detection capabilities of LVLMs, incorporating both positive and negative samples.
    \item Comprehensive analysis of leading LVLMs, including GPT-4o, Claude 3.5, Vila 2.7b and Qwen-VL, across multiple experimental settings, assessing their strengths and weaknesses in copyright infringement detection.
    \item Identification of failure cases and potential solutions, highlighting key challenges in current LVLM-based detection approaches and proposing improvements to enhance accuracy and robustness.
    \item By systematically investigating the role of LVLMs in copyright detection, our study provides valuable insights into the potential and limitations of AI-driven content moderation tools. Our findings underscore the need for continued research in this space to develop more reliable, ethically responsible, and legally compliant AI models capable of safeguarding intellectual property rights in the digital age.
\end{itemize}

\section{Related Work}

\subsection{Vision Language Models}
In recent years, Vision-Language Models (VLMs) have significantly advanced the integration of visual and textual data, leading to more sophisticated AI applications. A notable example is CLIP,  which employs contrastive learning to align images and text in a shared latent space, enabling zero-shot image classification and cross-modal retrieval \citep{pmlr-v139-radford21a}. Building upon such foundations, Large Vision Language Models (LVLMs) like GPT-4 have extended capabilities to process both textual and visual inputs, enhancing tasks such as image description and visual question answering \citep{gpt4v}. Similarly, Anthropic's Claude 3.5 has been developed to handle multimodal inputs, contributing to advancements in understanding and generating content across different modalities \citep{Claude3.5}. Further contributions include LLaVA, which integrates visual features into language models to improve visual reasoning \citep{liu2024visual}, and Qwen-VL, which supports multilingual conversations and end-to-end text recognition in images \citep{bai2023qwen}. Additionally, DeepSeek-VL2 has been recognized for its performance in visual understanding benchmarks, demonstrating the rapid progress in this field \citep{wu2024deepseek}. Collectively, these models represent significant contributions in combining image and text, paving the way for comprehensive AI systems. 
%CLIP \citep{pmlr-v139-radford21a}; GPT4 series \citep{gpt4v}; Claude 3.5 \citep{Claude3.5}; LLaVA \citep{liu2024visual}; Qwen-VL \citep{bai2023qwen}; DeepSeekVL2 \cite{wu2024deepseek}.

\subsection{Copyright Issues Related to Generative Models.}
The rapid advancement of generative AI enables the creation of text and images that closely mimic human-authored works, leading to significant legal and ethical concerns regarding potential infringements of intellectual property rights \cite{zirpoli2023generative, dzuong2024uncertain, sag2023copyright, polandgenerative}. A key contributing factor is that visual generative models may memorize portions of their training data, resulting in outputs that inadvertently reproduce IP-protected content \cite{carlini2023extracting, somepalli2023diffusion, gu2023memorization}. To mitigate IP infringement, two primary approaches have emerged:

\begin{itemize}
    \item Reducing Memorization During Training: Implementing differential privacy \cite{dwork2014algorithmic} techniques during the training of generative models can help minimize the retention of specific data points, thereby reducing the risk of reproducing protected content \cite{abadi2016deep, chen2022dpgen, dockhorn2022differentially}.
    \item Prompt Engineering: Employing strategies such as negative prompts during the inference phase can exclude undesired concepts or elements from the generated output\cite{wang2024evaluating, he2024fantastic}, or optimizing unsafe prompts \cite{chin2023prompting4debugging, rando2022red}, thereby avoiding the inclusion of IP-protected material.
\end{itemize}

Despite the widespread copyright concerns surrounding generative AI and the numerous IP mitigation approaches recently proposed, the issue of benchmarking VLM IP infringement detection remains largely underexplored. As a result, a primary focus of our paper is to address the capabilities of VLMs in detecting and mitigating IP infringement.

\subsection{In-context Learning}
In-context learning, as discussed in \cite{mann2020language, dong2022survey}, is a paradigm where large language models (LLMs) perform tasks by conditioning on a prompt that includes a few examples, enabling them to adapt to new tasks without explicit parameter updates. An in-context learning prompt generally includes two main parts: demonstrations and a new query. Demonstrations consist of several question-answer examples, each providing a full question along with its corresponding answer. The new query is a fresh question presented to the model for response. What's more, recent studies \cite{zhou2024visual,monajatipoor2023metavl,li2024visual, zhang2023makes} have demonstrated that vision-language models (VLMs) can also effectively facilitate in-context learning: Given a few images, or masks as examples, the VLMs could perform segmentation, classification, and visual question answering (VQA) \cite{antol2015vqa} in effective ways. 

% \section{Mitigation Approaches}

\section{Evaluation Benchmark Dataset}
\label{Sec 3}
%\subsection{Overview}
To evaluate the intellectual property (IP) infringement detection capabilities of LVLMs, the availability of comprehensive benchmark datasets is crucial. However, the lack of such datasets, especially those containing ambiguous negative IP infringement samples, poses a significant challenge for researchers aiming to thoroughly assess the detection abilities of different LVLMs. 

To bridge this gap, we have introduced a benchmark dataset, which includes images of widely recognized IP characters to support VLM evaluations. Our dataset curation and selection adhere to the following principles:  \ding{172} \textbf{IP Renown}: To effectively evaluate the ability of LVLMs to infringe on intellectual property (IP), our dataset includes both positive and negative samples derived from well-known, IP-protected figures.  \ding{173} \textbf{Diversity}: A comprehensive assessment of LVLM performance requires diverse image generation. Thus, we ensure a wide range of IP characters are included, with each character type synthesized using multiple generation techniques.  \ding{174} \textbf{Reproducibility}: Our dataset is built using open-source generative models with well-defined prompts, allowing for easy replication by other entities.

As a result, we focused on five iconic figures: Iron-Man, Batman, Spider-Man, Superman, and Super Mario. For each character, we gathered 200 images, ensuring a balanced distribution of positive and negative samples, all of which were meticulously labeled through human evaluation (refer to Table \ref{table1}). Additionally, to enhance the diversity of the dataset, we employed various methods to source images, such as collecting outputs from different generative models and utilizing diverse text prompt techniques, as elaborated below.

\begin{figure*}[ht]
    \centering
    \includegraphics[width=0.16\linewidth]{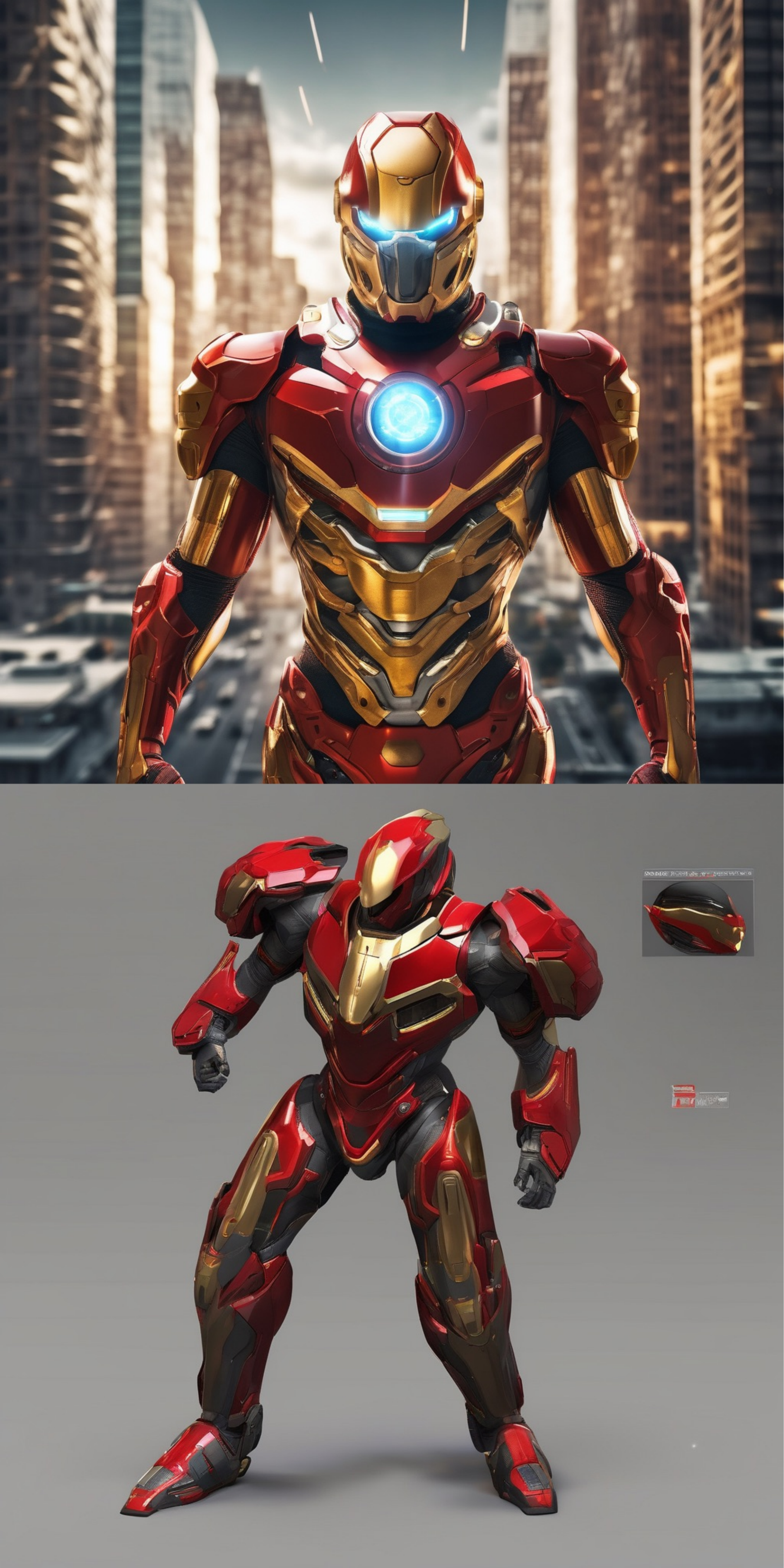}
    \includegraphics[width=0.16\linewidth]{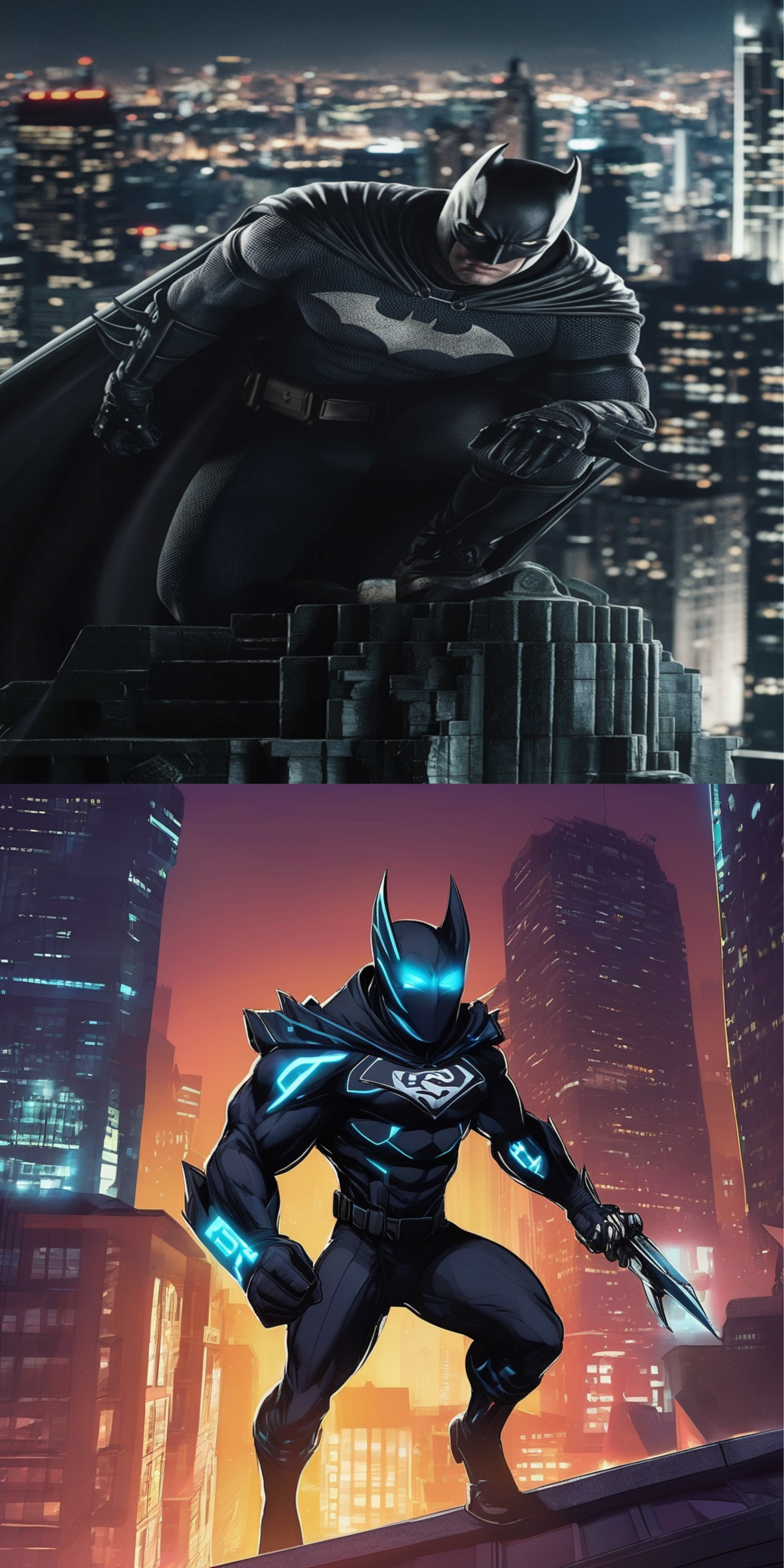}
    \includegraphics[width=0.16\linewidth]{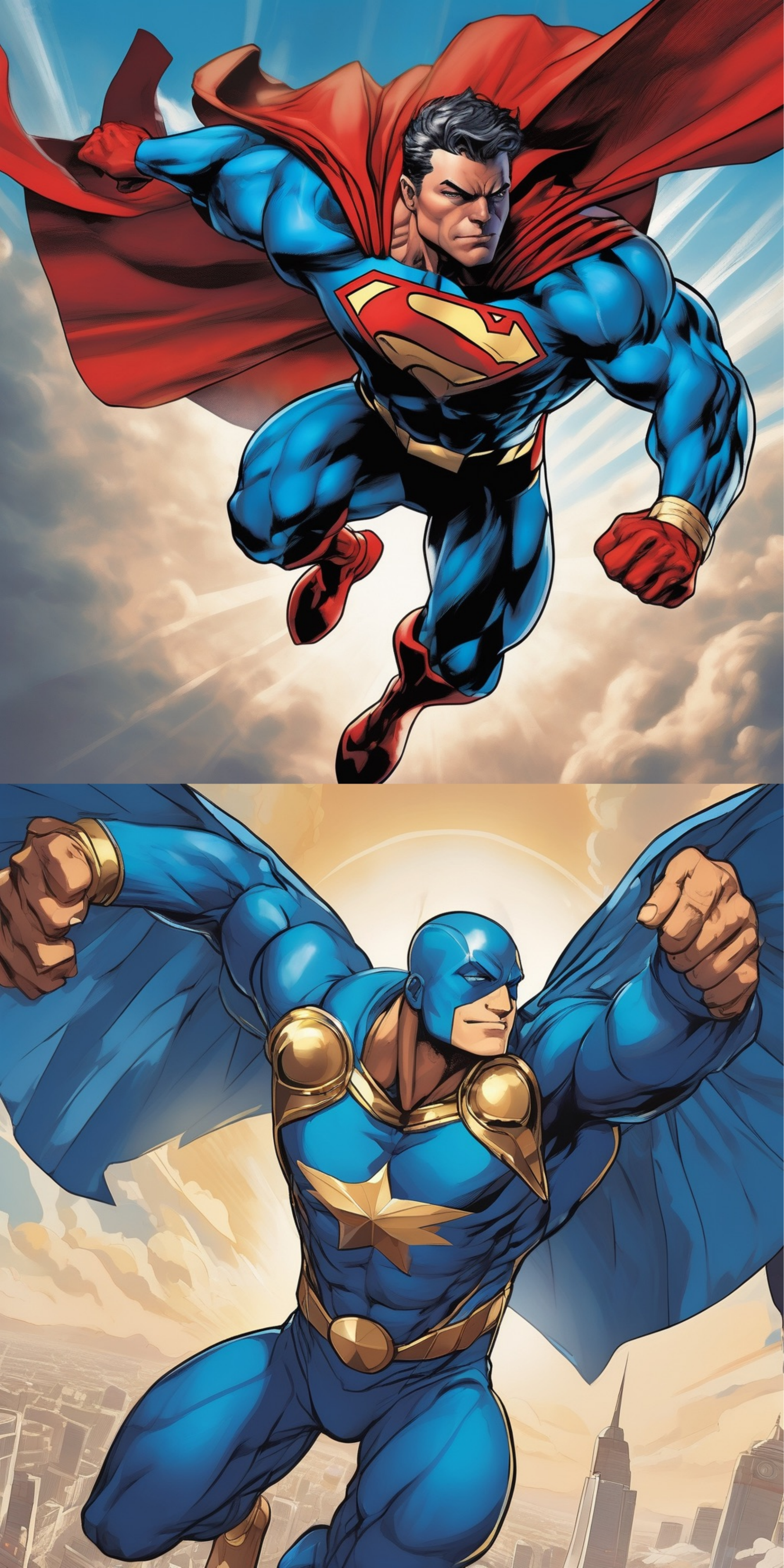}
    \includegraphics[width=0.16\linewidth]{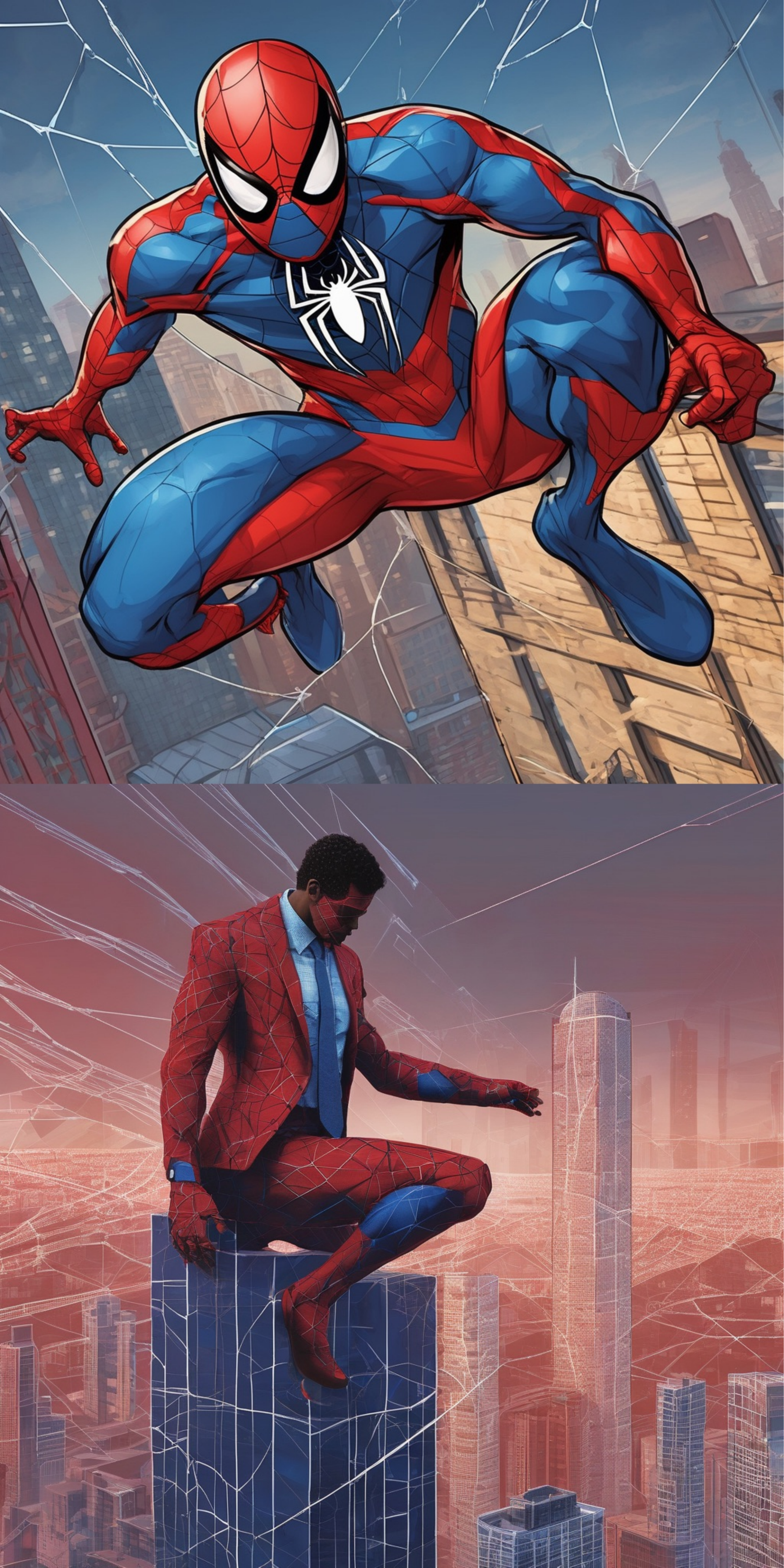}
    \includegraphics[width=0.16\linewidth]{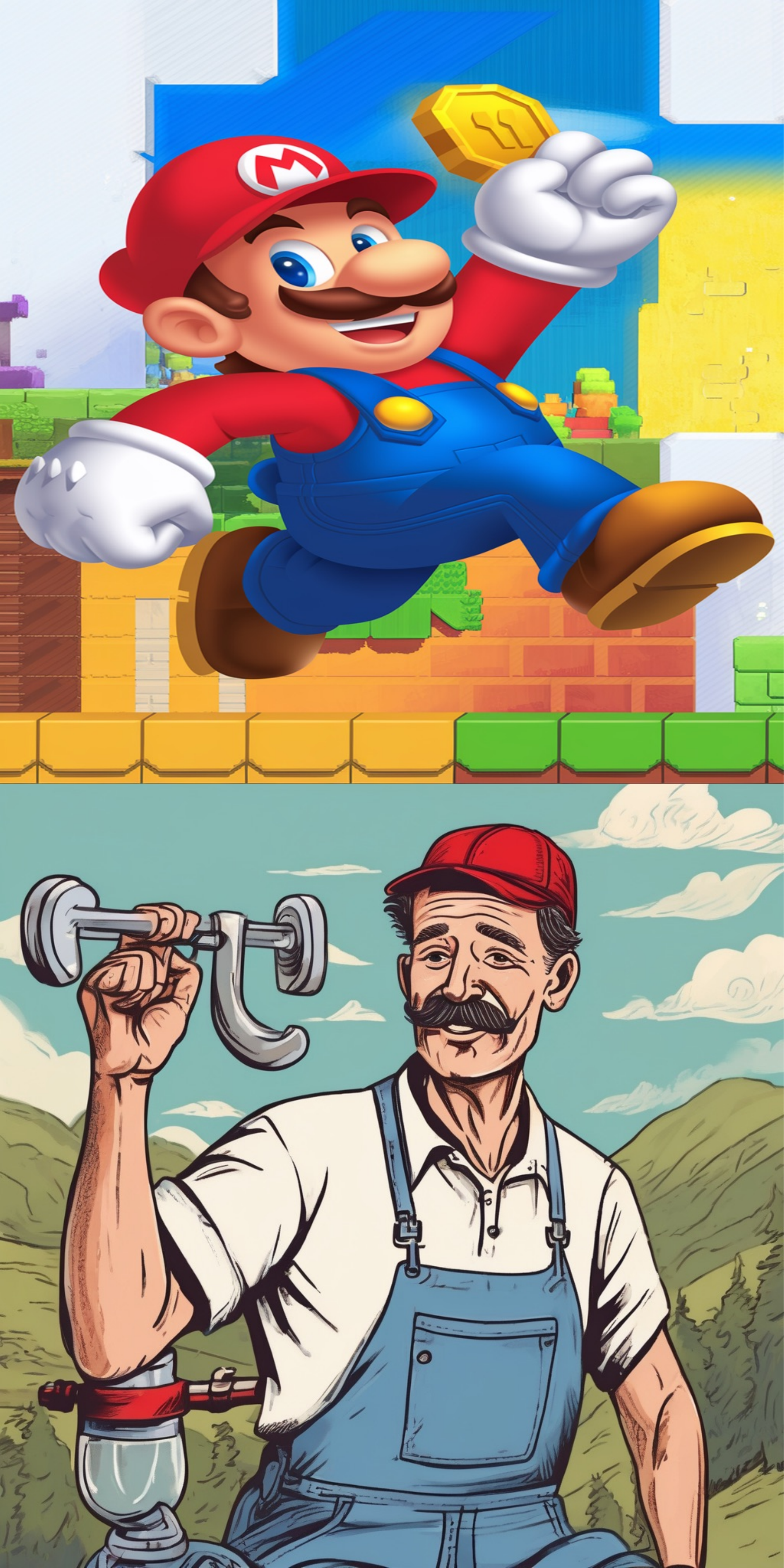}
    \caption{\textbf{Overview of positive and negative samples in the dataset.} First row: positive samples. Second row: Negative samples. Copyright protected characters from left to right: Iron-Man, Batman, Superman, Spider-man and Super-Mario.}
    \label{fig:1}
\end{figure*}

\begin{table}[h]
    \setlength{\tabcolsep}{1.5pt}
    \centering
    \begin{footnotesize}
    \begin{tabular}{lccccc}
    \toprule
       Class  & SpiderMan & BatMan & IronMan & SuperMan & SuperMario\\
    \midrule
       Ratio  & 0.41 & 0.51 & 0.44 & 0.53 & 0.64\\
    \bottomrule
    \end{tabular}
    \caption{Dataset Overview. "The ratio" indicates the proportion between positive samples and negative samples, where each class contains 200 image samples.}
    \label{table1}
    \end{footnotesize}
\end{table}

\subsection{Collecting Images}
In this section, we outline the principles behind generating and curating benchmark images using Stable Diffusion XL (SDXL) \citep{podell2023sdxl}, Ideogram \citep{ideogramAI}, DALL-E \citep{betker2023improving}, and Stable Diffusion XL-PerpNeg. Notably, SDXL-PerpNeg is an adaptation of the SDXL model \citep{podell2023sdxl} incorporating the PerpNeg method \citep{armandpour2023re}, which effectively mitigates image-based IP infringement. A detailed discussion on the impact of SDXL-PerpNeg is provided in the appendix.

\subsubsection{Collecting Positive Samples}
In this section, we outline two methods for generating positive samples that infringe on copyright protection laws. For this purpose, we selected three widely-used generative AI models: Stable Diffusion XL (generated 40\% of the image samples) \citep{podell2023sdxl}, Ideogram (generated 40\% of the image samples) \citep{ideogramAI}, and DALL-E (generated 20\% of the image samples) \citep{betker2023improving}.

\begin{figure}[ht]
        \centering
        \includegraphics[width=0.18\linewidth]{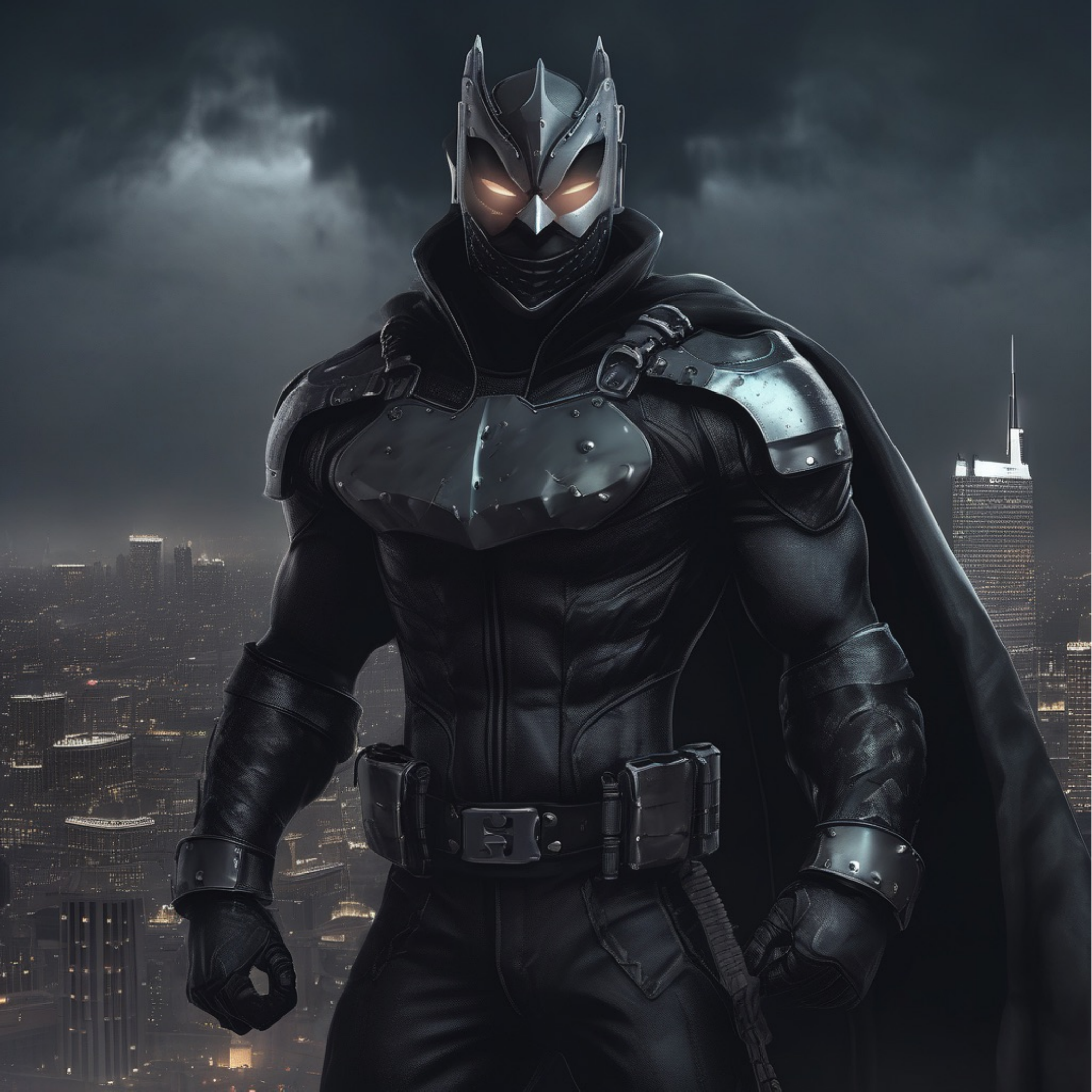}
        \includegraphics[width=0.18\linewidth]{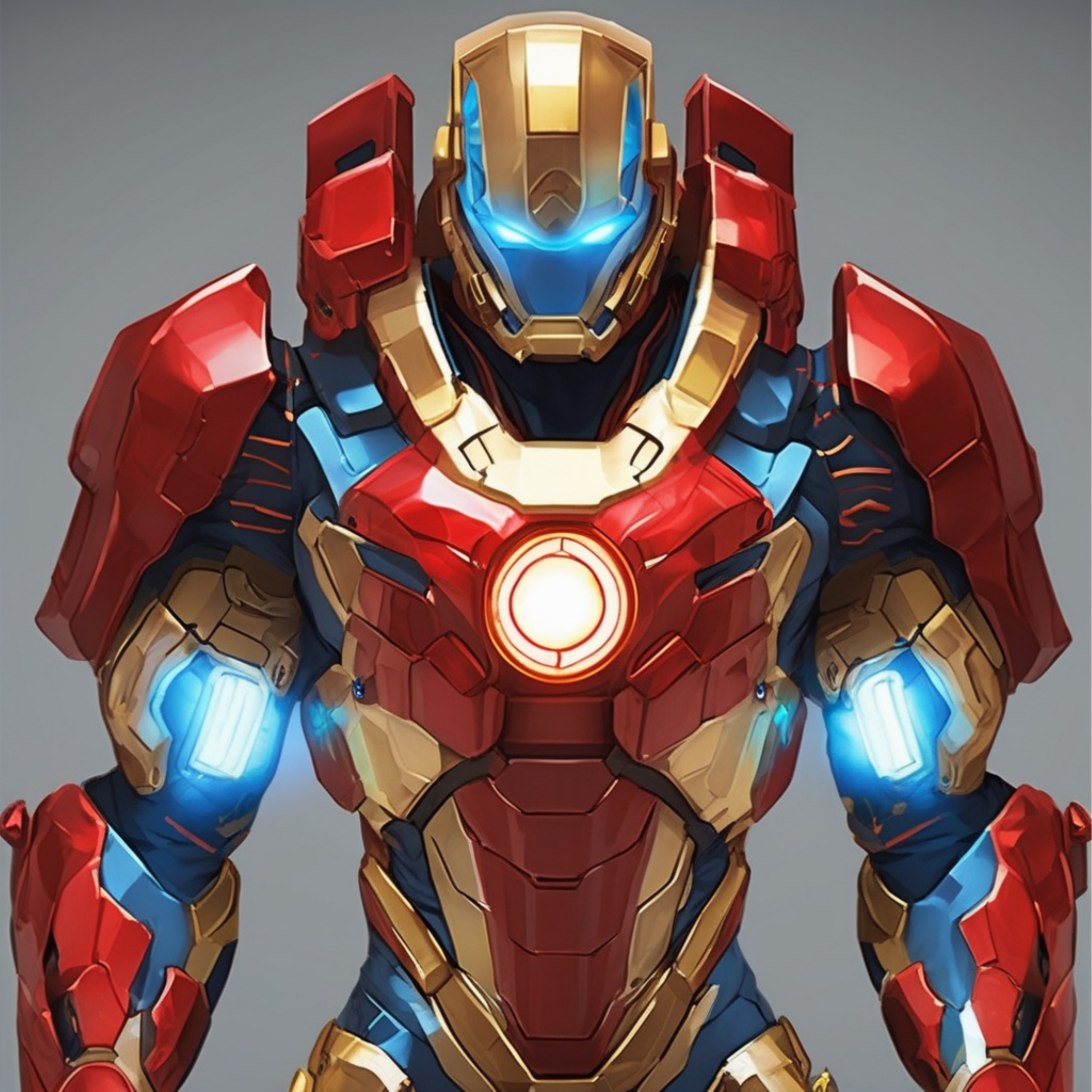}
        \includegraphics[width=0.18\linewidth]{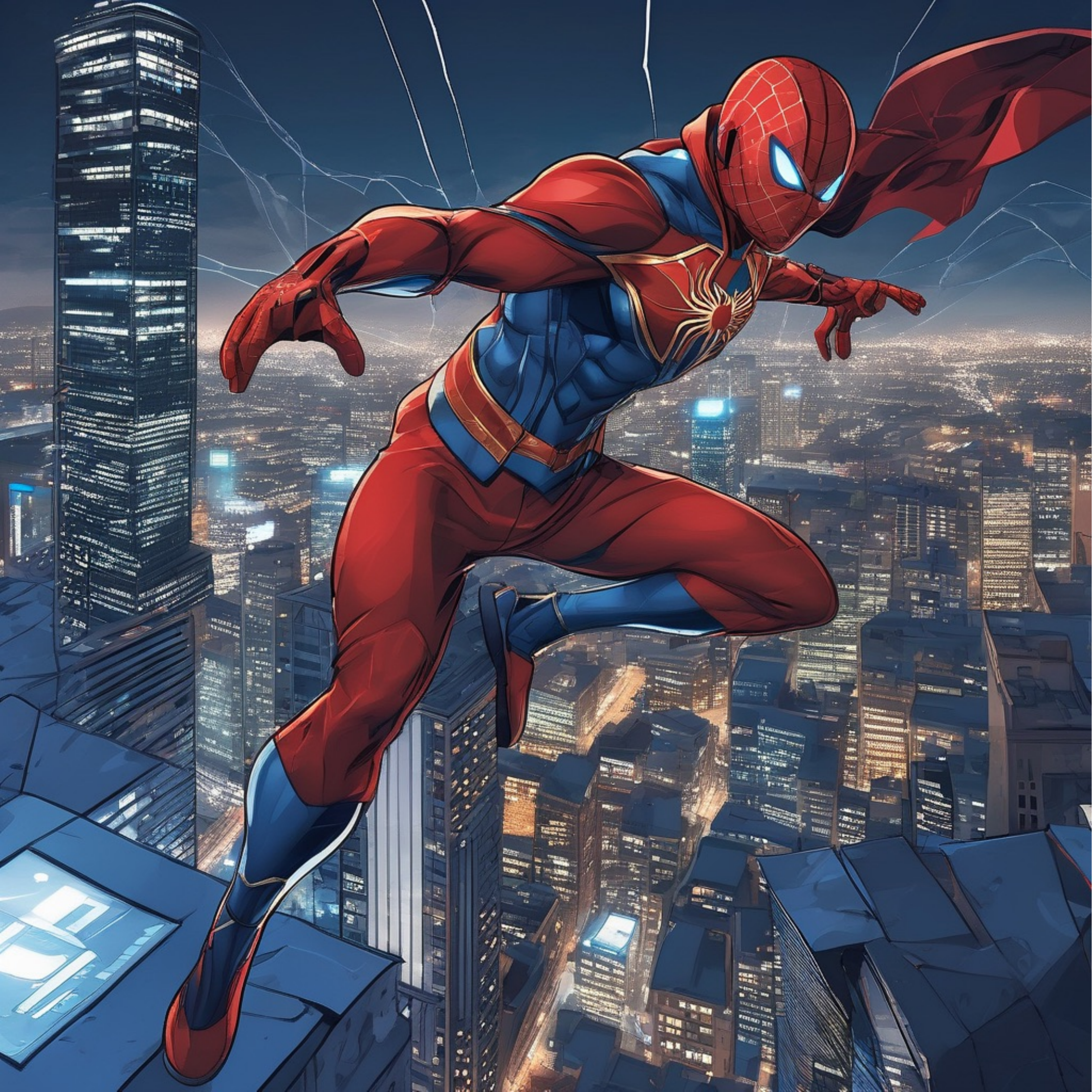}
        \includegraphics[width=0.18\linewidth]{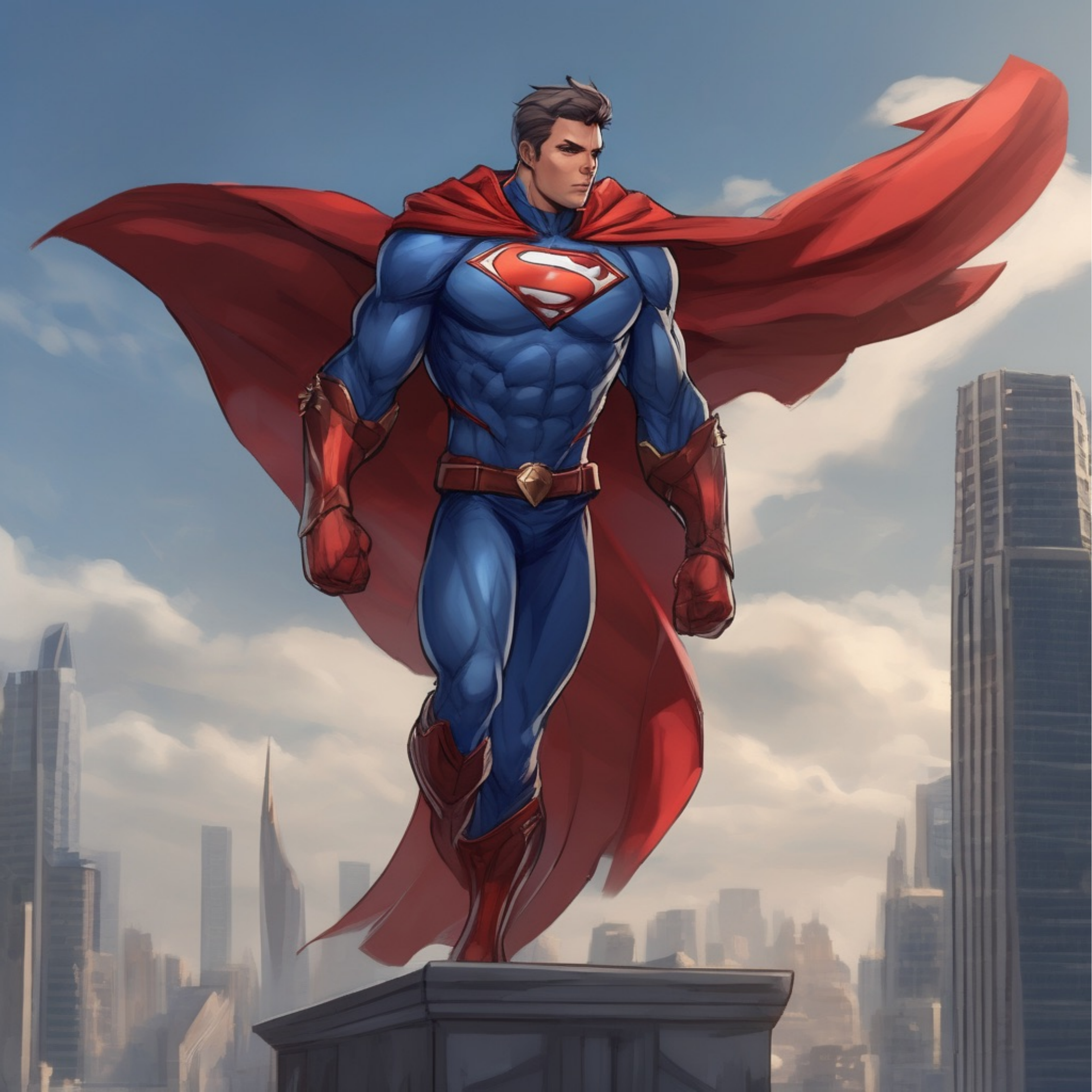}
        \includegraphics[width=0.18\linewidth]{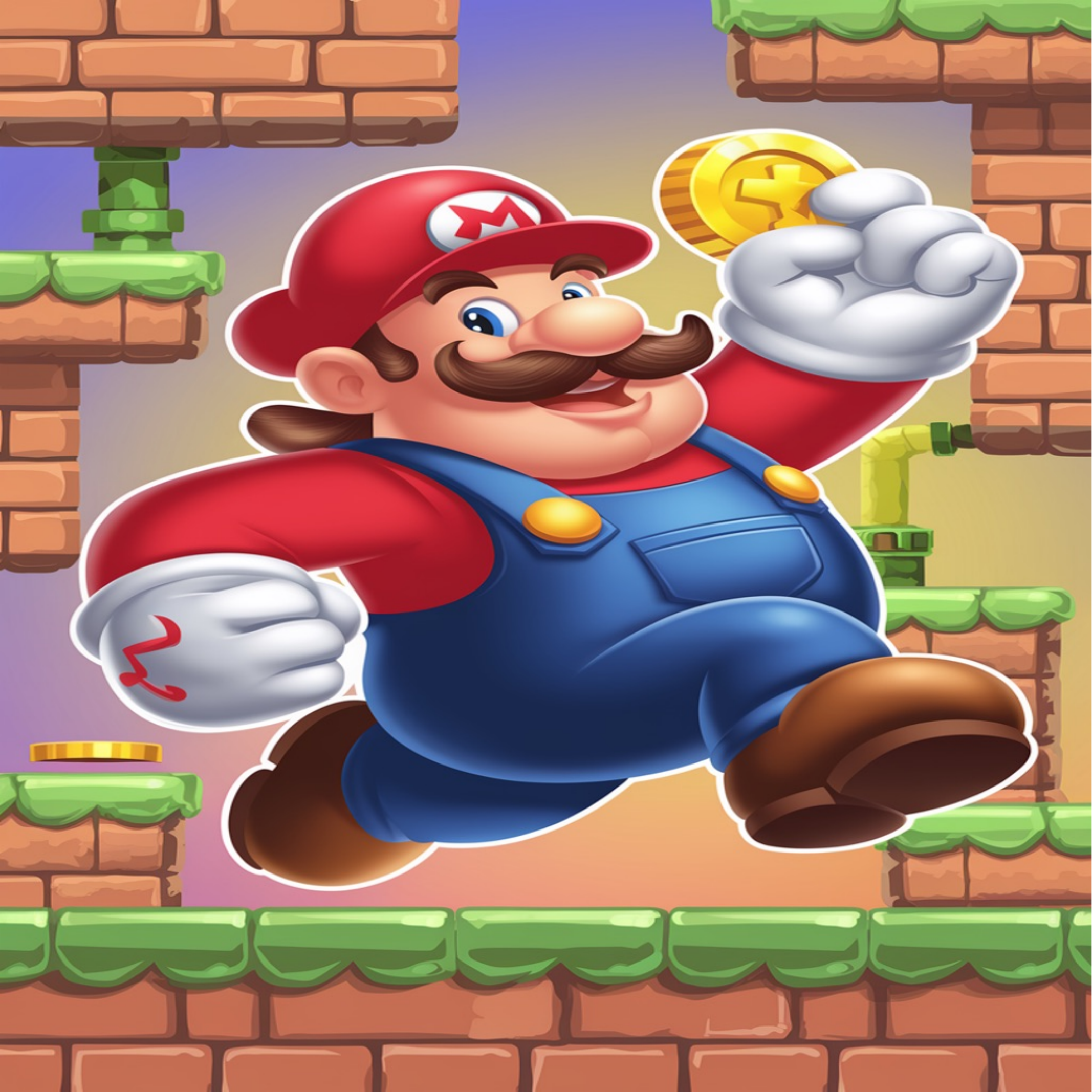}
        \caption{Generated positive samples with direct prompt from Stable Diffusion XL. Copyright protected characters from left to right: Iron-Man, Batman, Superman, Spider-man and Super-Mario.}
        \label{fig:2}
    \end{figure}

\noindent
\textbf{Generate IP Characters with Direct Prompt.} The simplest approach to generating positive samples involves using direct prompts with generative models, such as \textit{"Generate an image of <a character>."} This method typically produces images that closely resemble the IP-protected characters, as illustrated in Fig. \ref{fig:2}. Using this technique, we generated 40 images for each character class.

\noindent
\textbf{Generate with Descriptive Prompt.} Rewriting direct prompts that reference copyright-protected content into longer, more descriptive prompts, as explored by \citet{wang2024evaluating} and \citet{he2024fantastic}, can sometimes reduce the risk of IP infringement. However, this approach is not entirely effective in preventing outputs that closely resemble copyrighted characters \citep{he2024fantastic}, as rewritten prompts often retain a high degree of similarity to the original IP-associated names.  

\begin{figure}[ht]
        \centering
        \includegraphics[width=0.18\linewidth]{figures/Descriptive_prompt/batman.pdf}
        \includegraphics[width=0.18\linewidth]{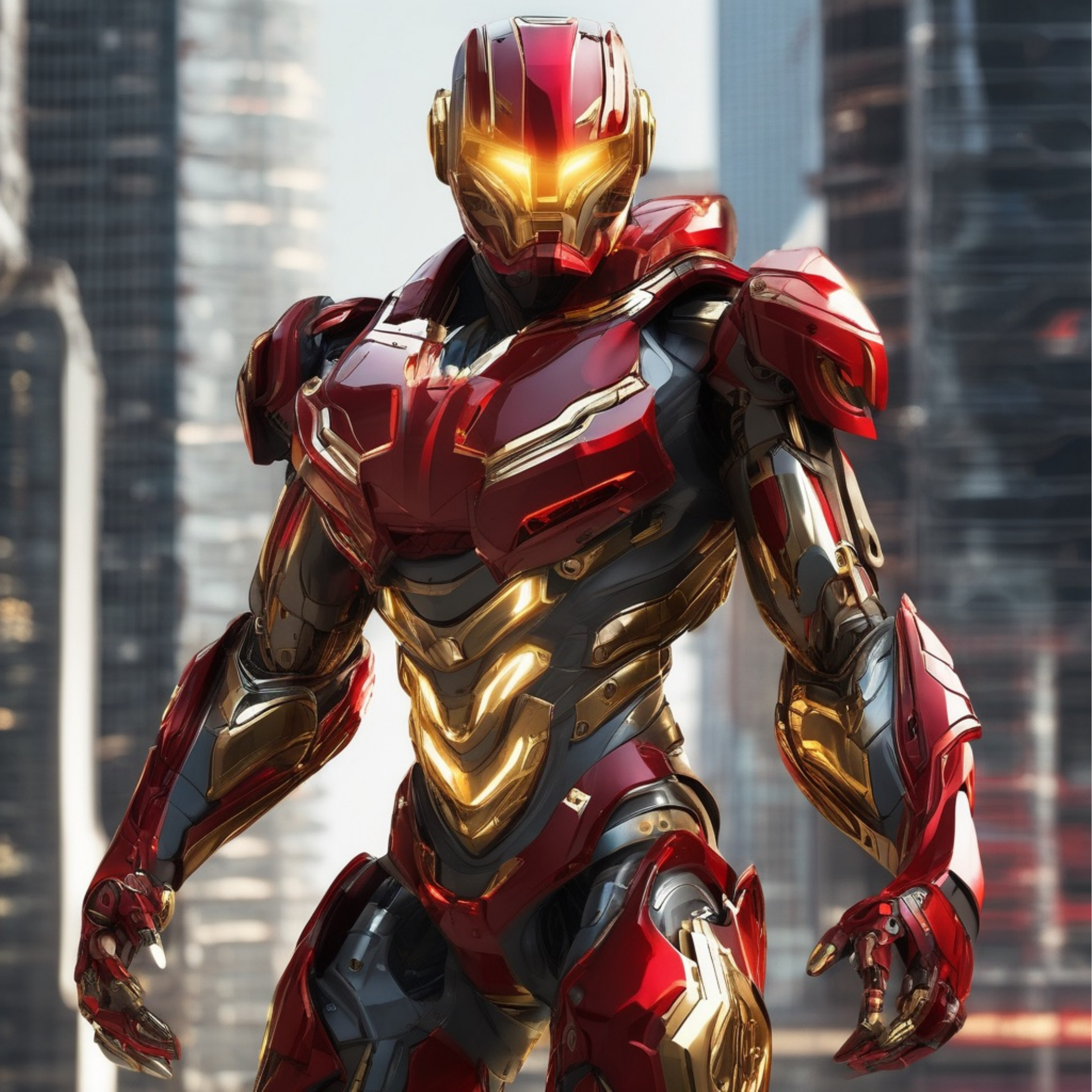}
        \includegraphics[width=0.18\linewidth]{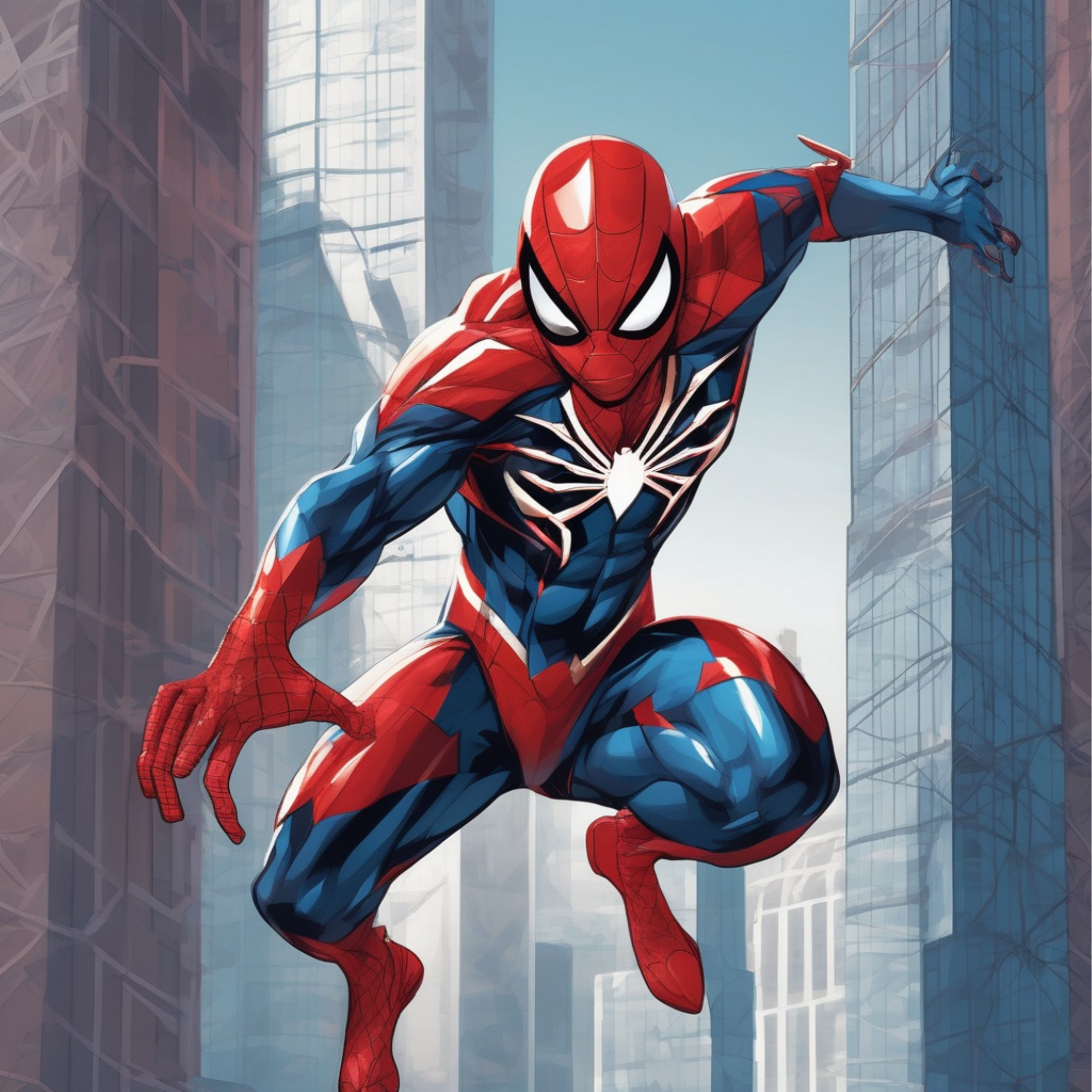}
        \includegraphics[width=0.18\linewidth]{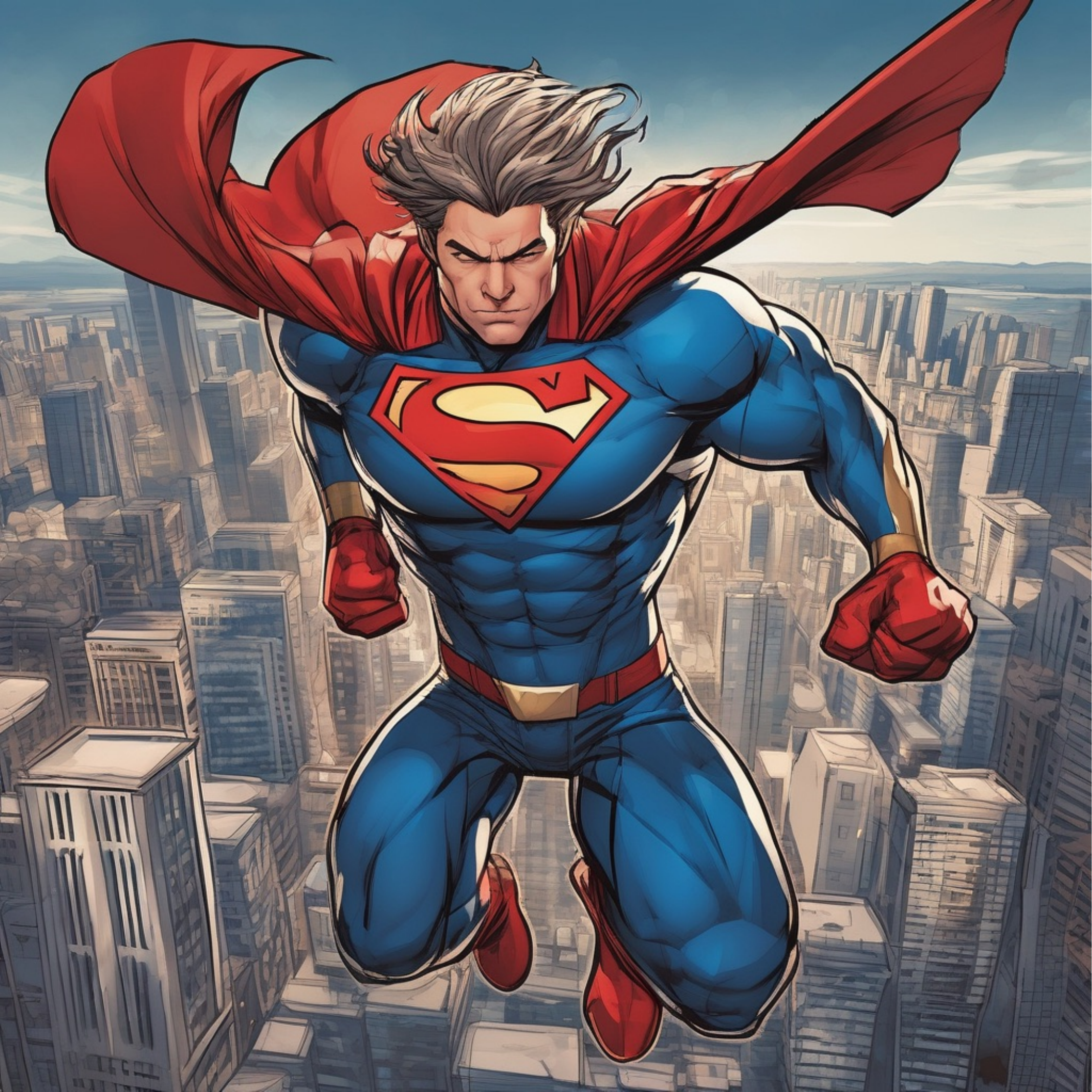}
        \includegraphics[width=0.18\linewidth]{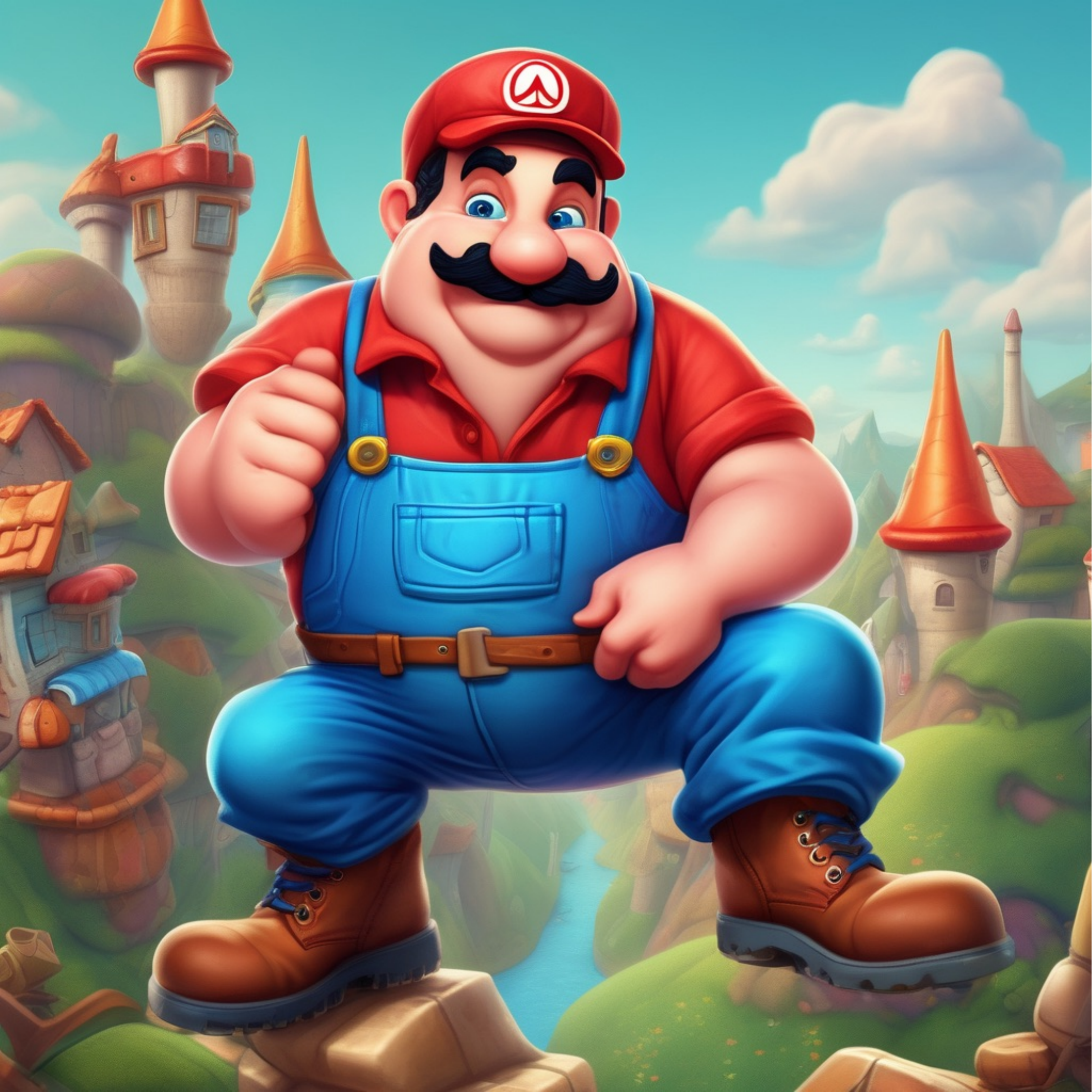}
        \caption{Generated positive samples with descriptive prompt (w/o character's name) from Ideogram AI. Copyright protected characters from left to right: Iron-Man, Batman, Superman, Spider-man and Super-Mario.}
        \label{fig:3}
    \end{figure}

To enhance the diversity of our dataset, we first generate images using descriptive prompts and then apply a human evaluation process to filter out most positive samples, as detailed in Section \ref{label img}. We use GPT-4o \cite{gpt4v} here as its exceptional text generation capabilities. We construct descript prompt with the following guidance to GPT-4o:
    \begin{itemize}
        \item Creating a prompt that describes a character similar to <Target
    Character>. This prompt should enable text-to-image AI models
    to generate images without directly mentioning the name of the
    <Target Character>.
    \end{itemize}
Finally, we curate the selected positive images, as illustrated in Fig. \ref{fig:3}.

\subsubsection{Collecting Negative Samples}
\begin{figure}[ht]
        \centering
        \includegraphics[width=0.18\linewidth]{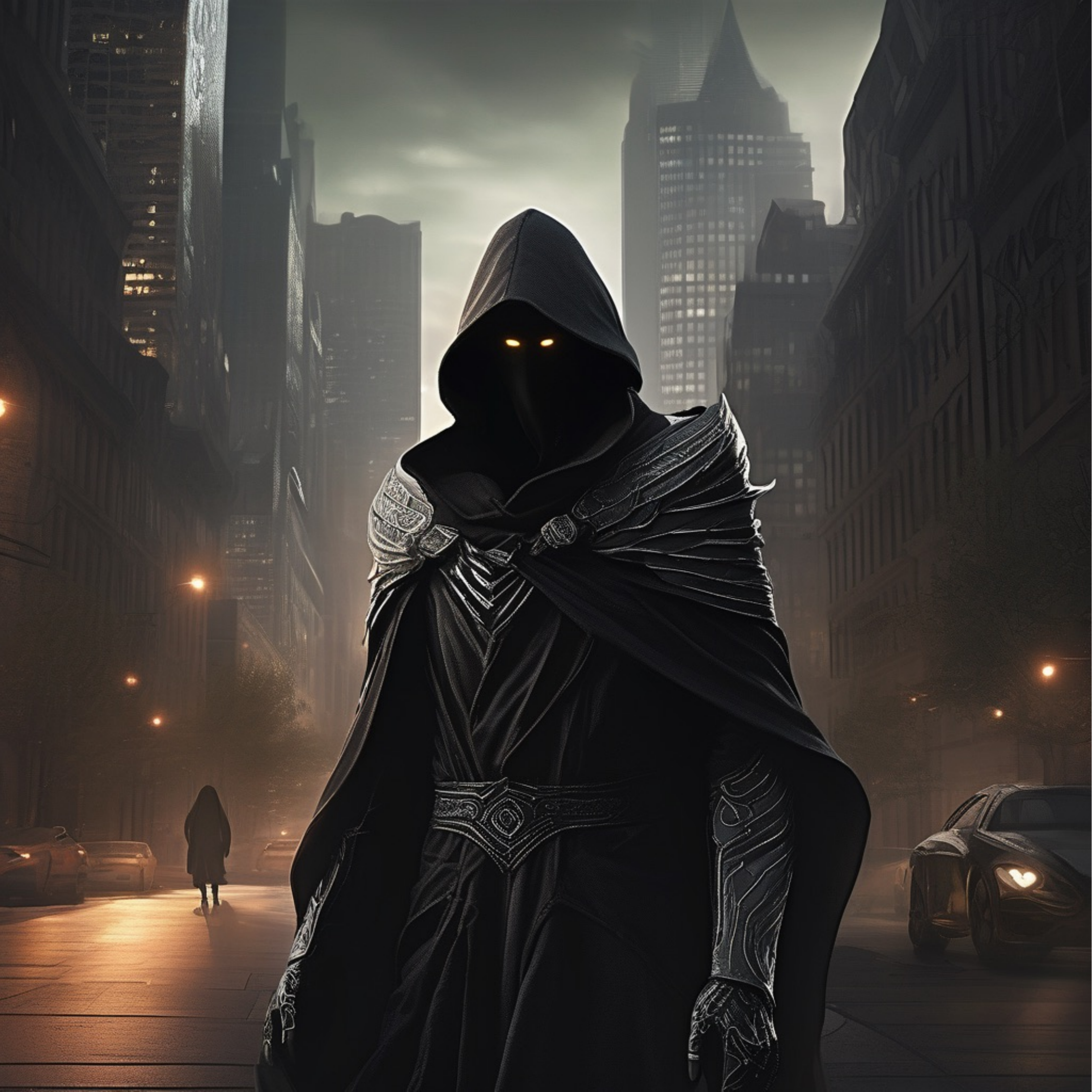}
        \includegraphics[width=0.18\linewidth]{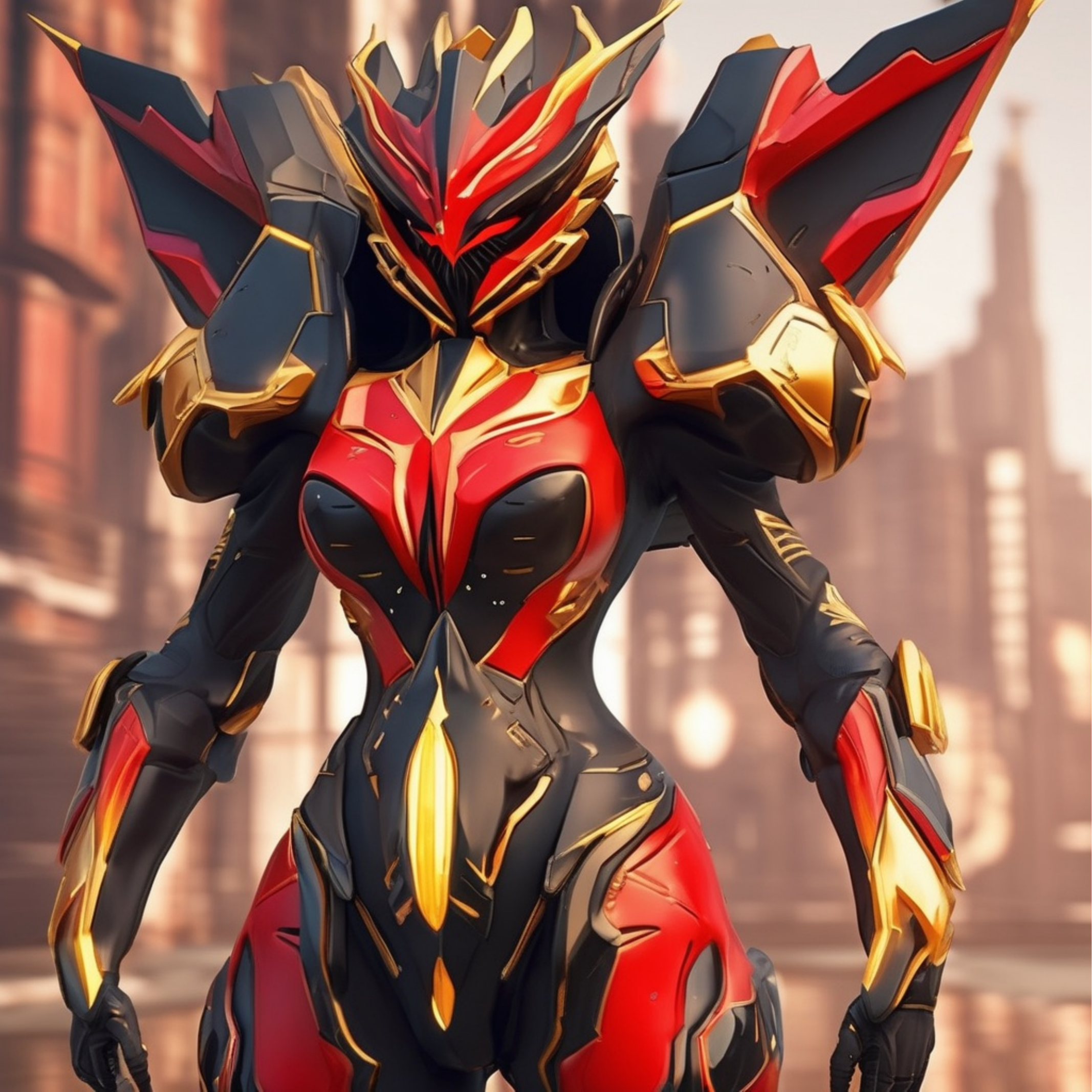}
        \includegraphics[width=0.18\linewidth]{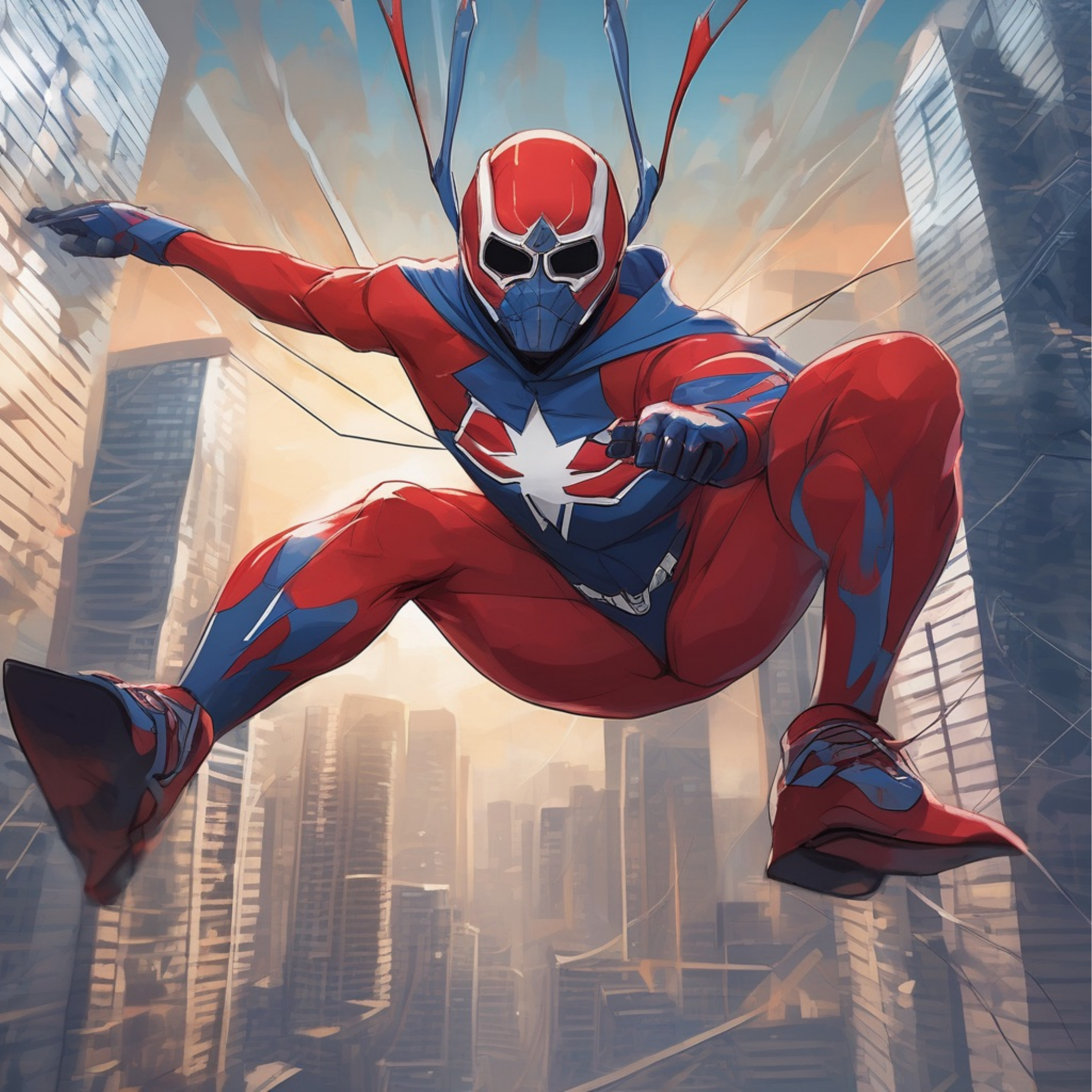}
        \includegraphics[width=0.18\linewidth]{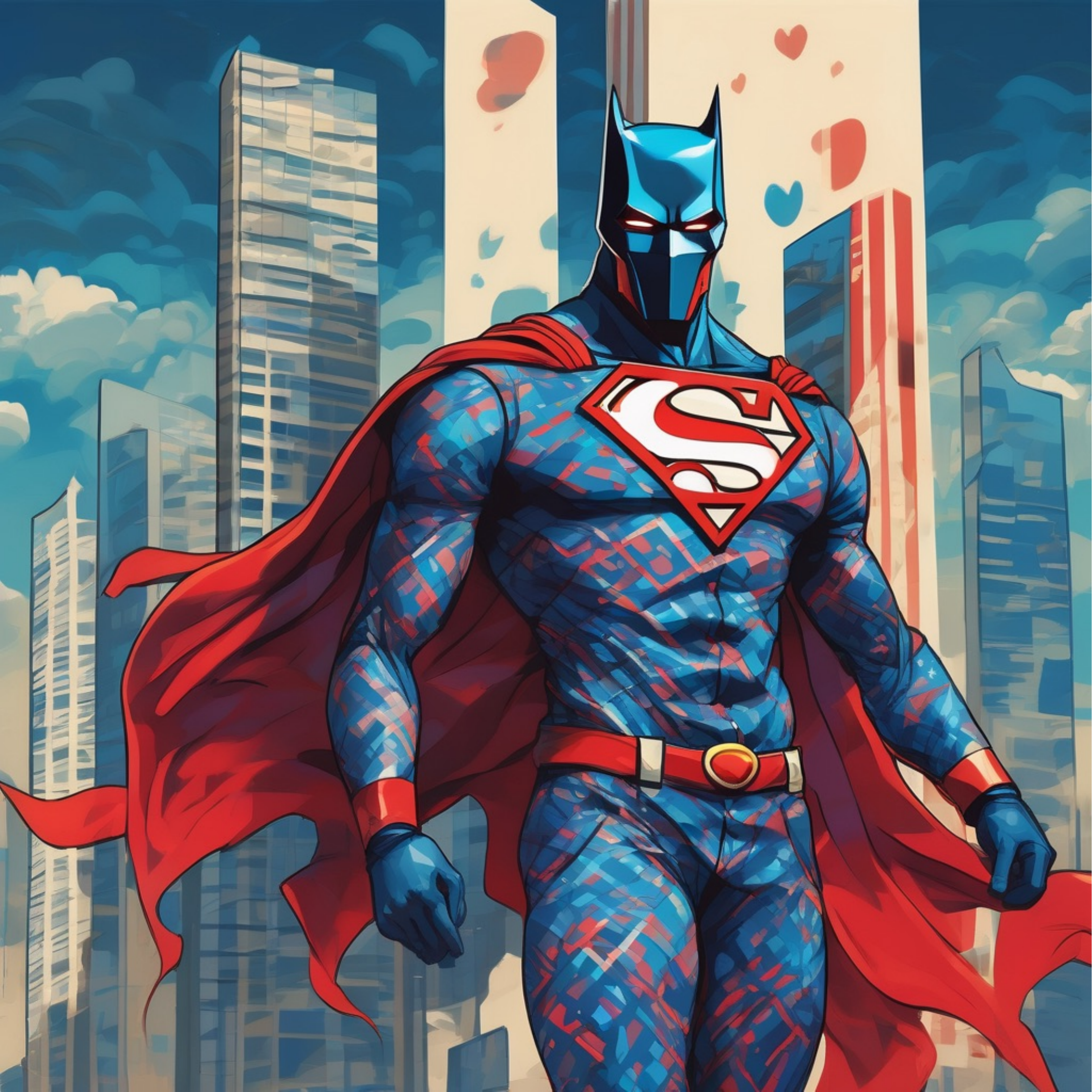}
        \includegraphics[width=0.18\linewidth]{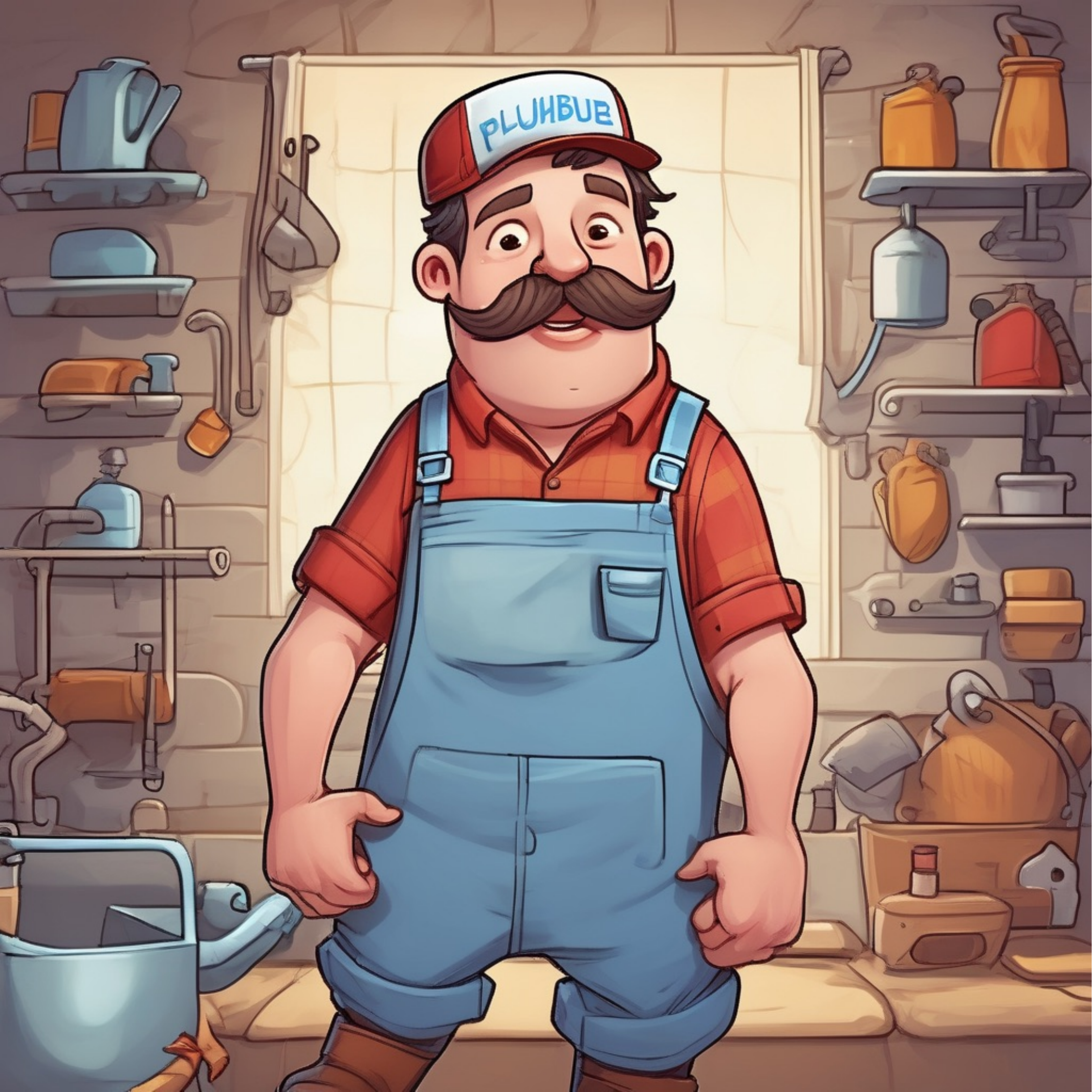}
        \caption{Generated negative samples with character name as negative prompts from DALL-E. Copyright protected characters from left to right: Iron-Man, Batman, Superman, Spider-man and Super-Mario.}
        \label{fig:4}
    \end{figure}

    \textbf{Generate with Plain Negative Prompts.} Negative prompts are commonly utilized in deploying diffusion models to enable users to exclude unwanted concepts or elements from the generated output. Incorporating the negative prompt through classifier-free guidance \cite{ho2022classifier}, the predicted noise gradient $\Tilde{\epsilon}_\theta(z_t, t, c)$ between timestamp $t$ and $t-1$ can be written as
    \begin{equation}
        \Tilde{\epsilon}_\theta(z_t, t, c, d) = \epsilon_\theta(z_t, t, c) - w(\epsilon_\theta(z_t, t, d)- \epsilon_\theta(z_t, t))
    \end{equation}
    where $z_t$ is the noise at timestamp $t$, $\epsilon_\theta$ is the UNet of diffusion models, $c$ is the non-negative text prompt, $d$ is the negative prompt, and $w$ is the scale parameter of the classifier-free guidance (we set $w$ to default value 7.5). By combining negative prompts, such as IP character names or commercial brands, the generative model can be guided away from synthesizing features or contents that resemble the infringing characters, as shown in Fig. \ref{fig:4}. 

\begin{figure}[ht]
        \centering
        \includegraphics[width=0.18\linewidth]{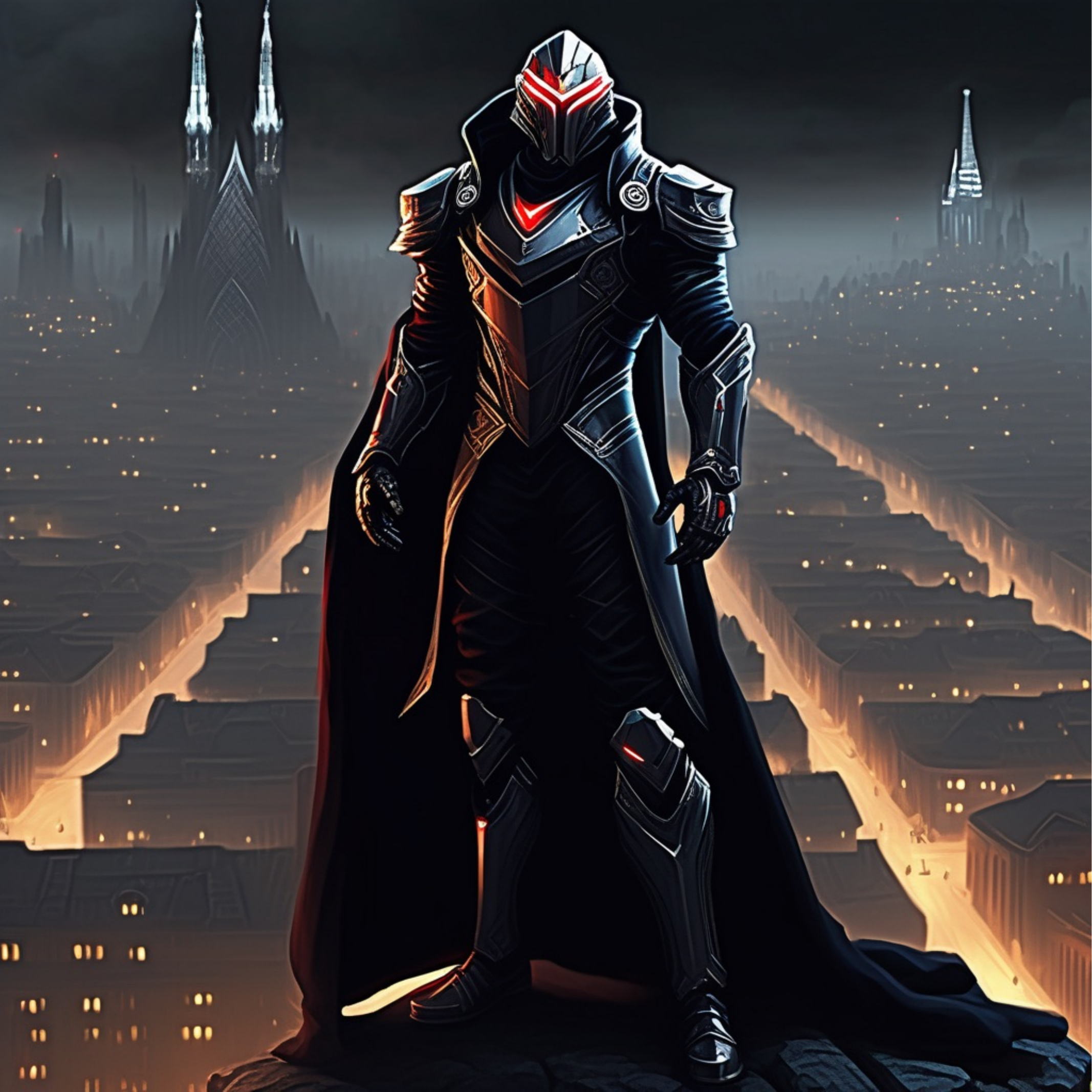}
        \includegraphics[width=0.18\linewidth]{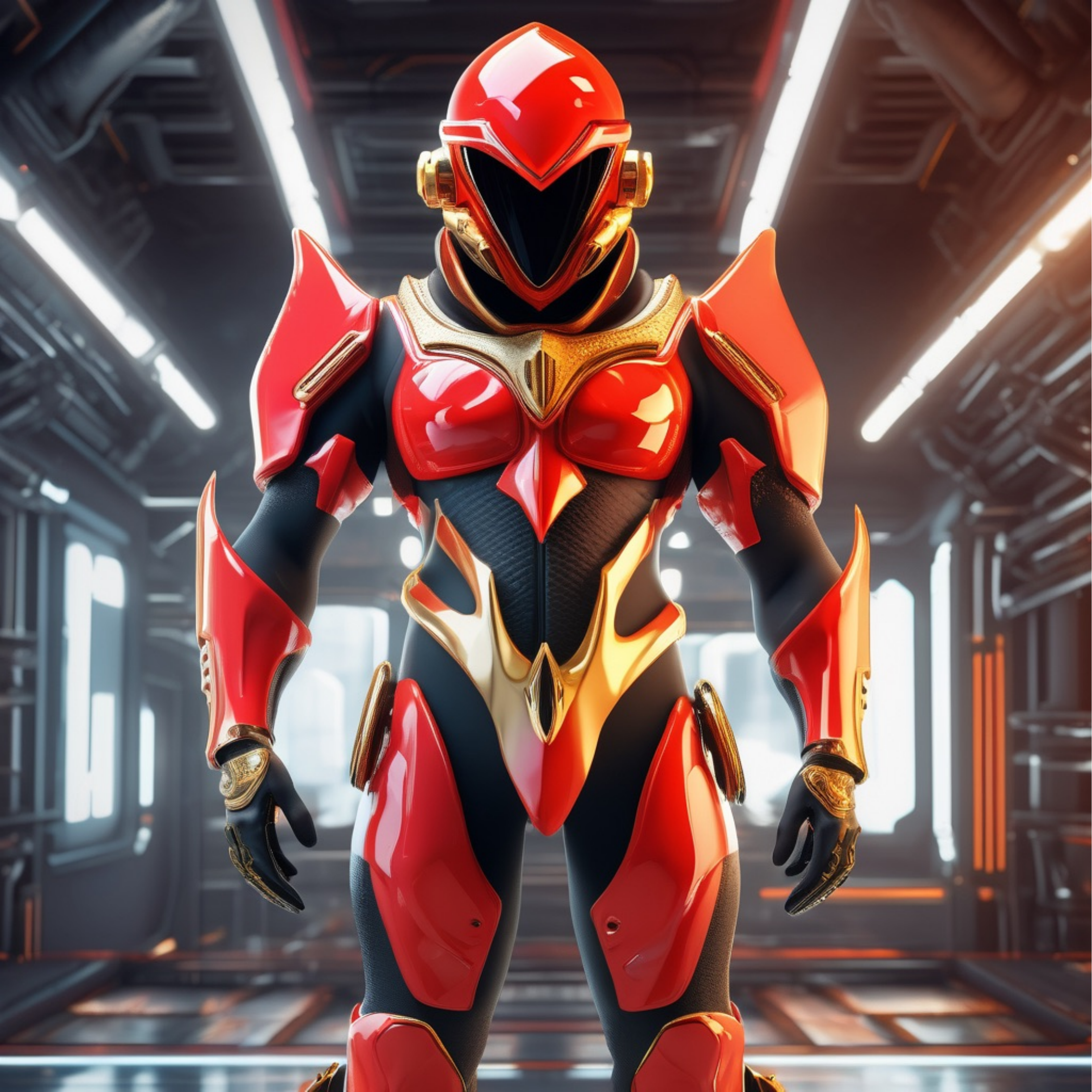}
        \includegraphics[width=0.18\linewidth]{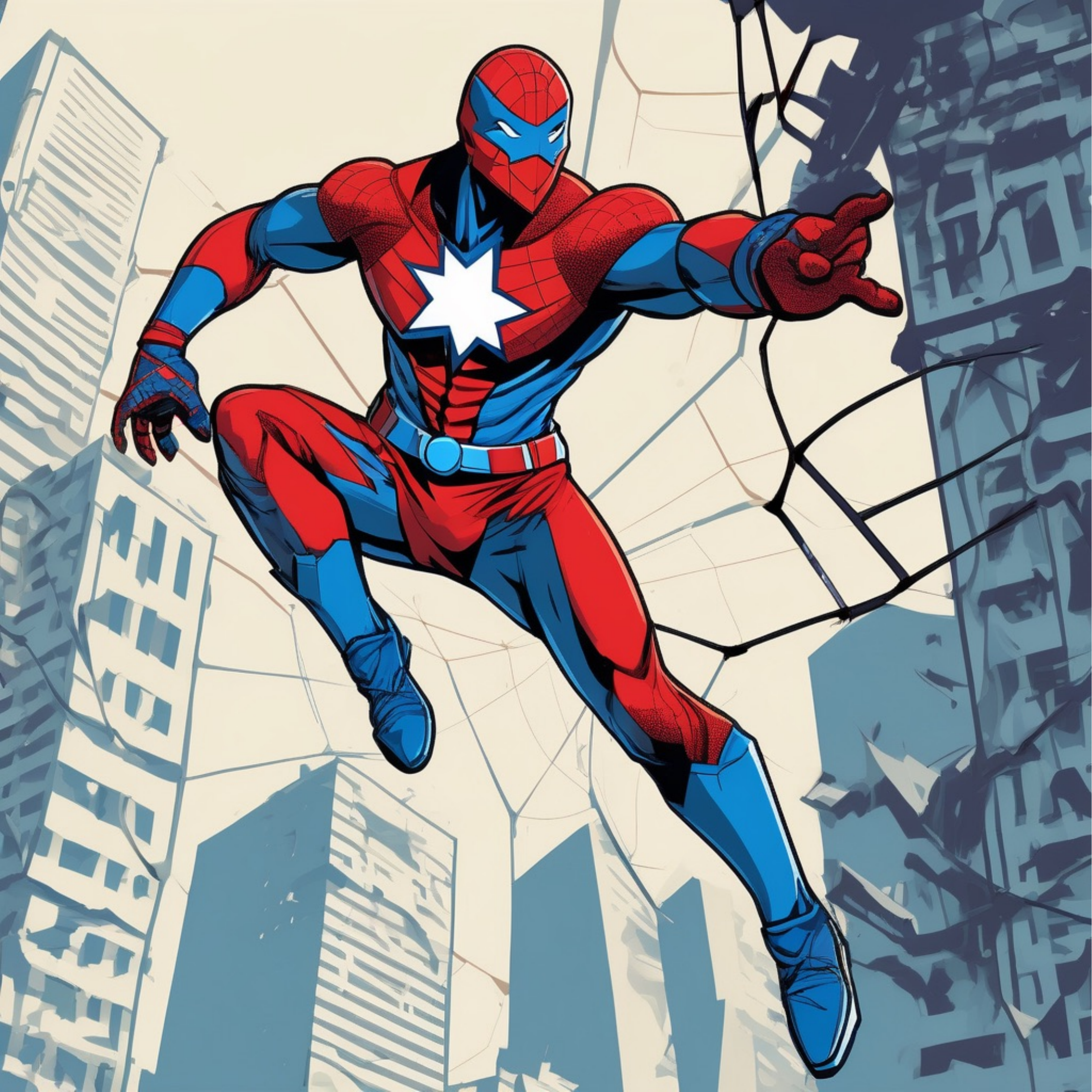}
        \includegraphics[width=0.18\linewidth]{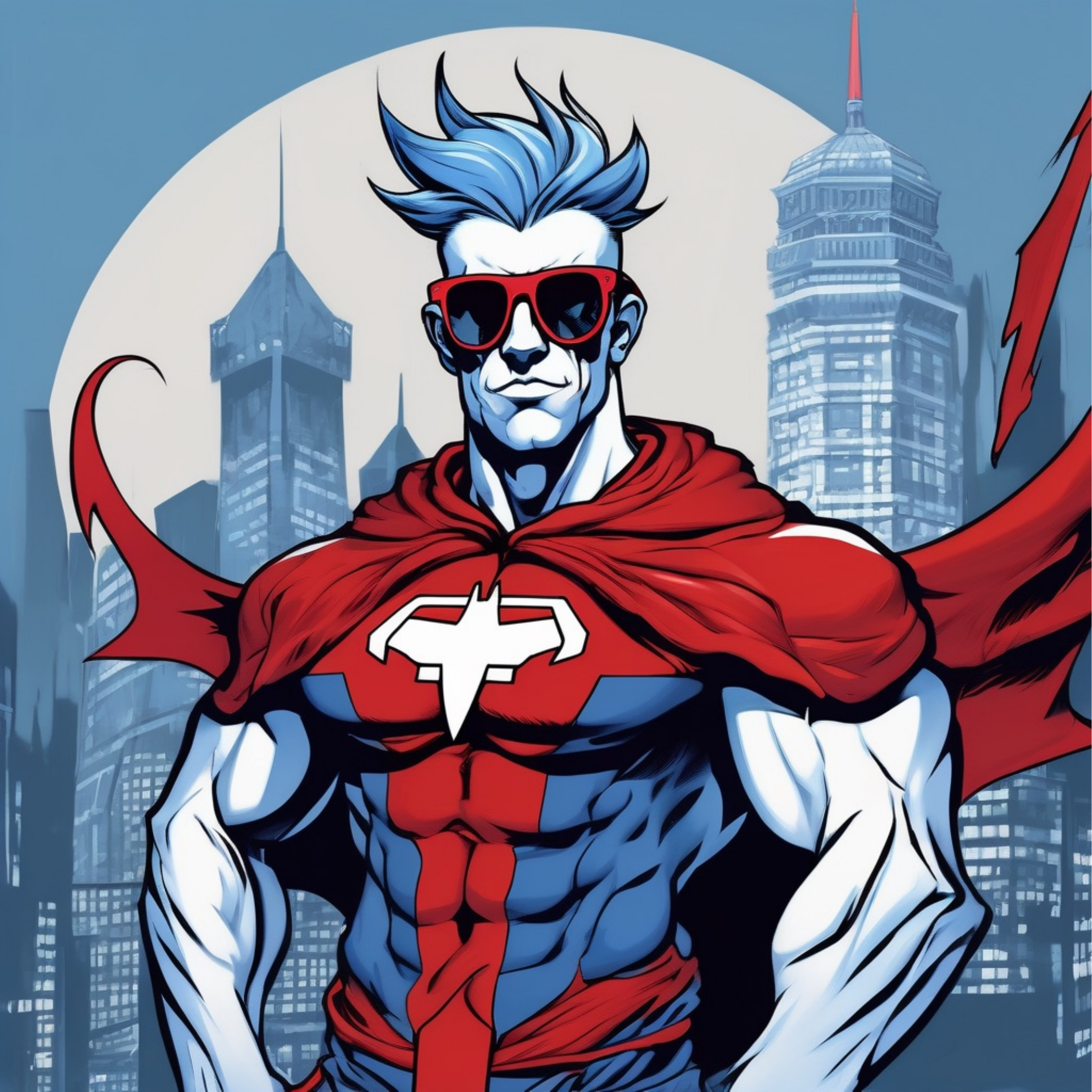}
        \includegraphics[width=0.18\linewidth]{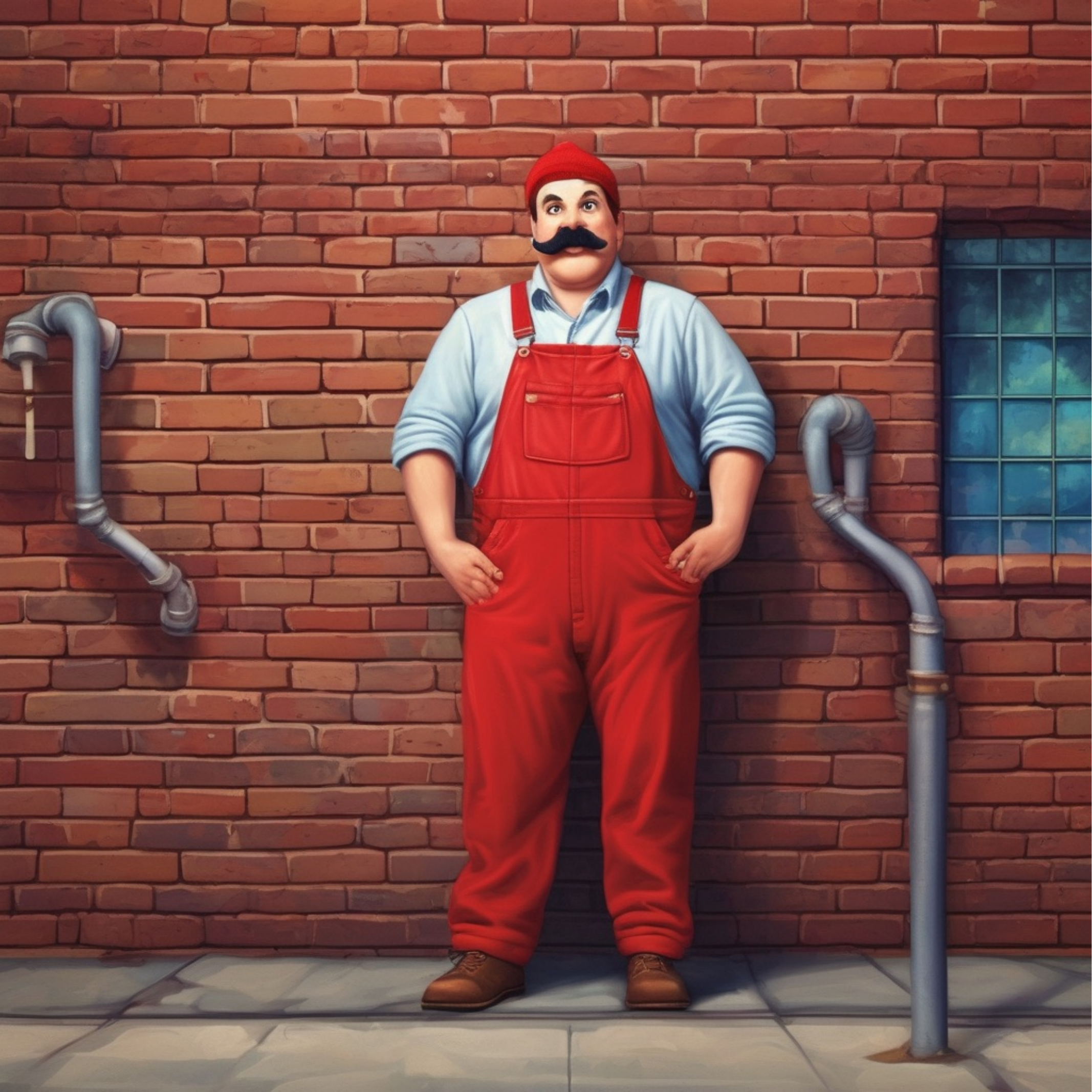}
        \caption{Generated negative samples with character name as negative prompts from Stable-Diffusion XL Perp-Neg. Copyright protected characters from left to right: Iron-Man, Batman, Superman, Spider-man and Super-Mario.}
        \label{fig:5}
    \end{figure}

    \noindent
    \textbf{Generate with Non-overlapped Negative Prompts (NNP).} When the input text prompt \( c \) and the negative prompt \( d \) have overlapping semantics, composing positive and negative prompts linearly could lead to undesired results, particularly in cases of concept negation. Perpendicular negative prompt, as discussed in \cite{armandpour2023re}, improved the overlap problem by applying the perpendicular gradient:
    \begin{multline*}
        \Tilde{\epsilon}_\theta^{Perp-Neg}(z_t, t, c_{pos}, d_i) = w_{pos}\epsilon_\theta^{pos} + \epsilon_\theta(z_t, t)\\ 
        - \sum_iw_i(\epsilon_\theta^{i}- \frac{\langle\epsilon_\theta^{pos}, \epsilon_\theta^{i}\rangle}{||\epsilon_\theta^{pos}||^2}\epsilon_\theta^{pos})
    \end{multline*}
    where negative prompt defined as $\epsilon_\theta^{i}= \epsilon_\theta(z_t, t, d_i)- \epsilon_\theta(z_t, t)$, with $w_{pos} > 0$ and $w_i > 0$ as the weight for each positive and negative prompt. Incorporating Perp-Neg with state-of-the-art generative model Stable Diffusion XL \citep{podell2023sdxl}, we can collect high-quality samples while getting rid of copyright infringement at the same time. As a result, we leverage Perp-Neg SDXL as a effective tool for generating negative samples, as shown in Fig. \ref{fig:5}.

    \noindent
    \textbf{Summary}. Collecting negative samples using negative prompt method, as shown in Algorithm \ref{alg:Perp_Neg_Prompt}, ensures that unwanted features associated with a negative prompt (e.g., a copyrighted character) are effectively suppressed while maintaining the desired aspects of the generated image. The second approach, using non-overlapping negative prompts, provides more precise control over feature removal.

\begin{algorithm}[!t]
\captionsetup{font=small}
\caption{Negative Samples Generation}
\label{alg:Perp_Neg_Prompt}
\begin{small} % Change \begin{small} to \begin{scriptsize}
\begin{algorithmic}[1]
\REQUIRE IP Character Name $D$, Pretrained Multi-modal Generative Model $\epsilon_\theta$, Negative Prompt Weight $w$.\\
\noindent{\bfseries Return} Negative Samples $\mathcal{I}$.
\STATE {Generate a descriptive prompt for the IP figure $C \leftarrow \epsilon_\theta(D)$}
\STATE {$z_T \leftarrow$ Random Noise Sample}
\IF {Generate with Plain Negative Prompt}
    \FOR{\(t\) \text{in} \([T,T-1,\dots,0]\)}
        \STATE {$\Tilde{\epsilon}_\theta(z_t, t, C, D) = \epsilon_\theta(z_t, t, C) - w(\epsilon_\theta(z_t, t, D)- \epsilon_\theta(z_t, t))$}
        \STATE {$z_{t-1} = \alpha_t \cdot z_{t} + \beta_t \cdot \Tilde{\epsilon}_{\theta} (z_t, t, C, D)$}
    \ENDFOR
    \STATE {$\mathcal{I} = z_0$}
\ENDIF
\IF{Generate with Non-overlapped Negative Prompts}
    \FOR {\(t\) \text{in} \([T,T-1,\dots,0]\)}
        \STATE {$\epsilon_\theta^{neg} = \epsilon_\theta(z_t, t, D)$}
        \STATE {$\epsilon_\theta^{pos} = \epsilon_\theta(z_t, t, C)$}
        \STATE {${\epsilon}_\theta^{Perp}(z_t, t, C, D) = w(\epsilon_\theta^{neg} - \frac{\langle\epsilon_\theta^{pos}, \epsilon_\theta^{neg}\rangle}{||\epsilon_\theta^{pos}||^2}\epsilon_\theta^{pos})$}
        \STATE {$\Tilde{\epsilon}_\theta(z_t, t, C, D) = {\epsilon}_\theta(z_t, t, C) + {\epsilon}_\theta(z_t, t)- {\epsilon}_\theta^{Perp}(z_t, t, C, D)$}
        \STATE {$z_{t-1} = \alpha_t \cdot z_{t} + \beta_t \cdot \Tilde{\epsilon}_{\theta} (z_t, t, C, D)$}
    \ENDFOR
    \STATE {$\mathcal{I} = z_0$}
\ENDIF
\STATE \textbf{return} $\mathcal{I}$
\end{algorithmic}
\end{small} % Change \end{small} to \end{scriptsize}
\end{algorithm}
    
% \textbf{important points: How to collect/generate negative samples?}
\subsection{Labeling Images}
After gathering the images, we employ human annotators to label the dataset samples. The process begins by selecting 50 images from the entire dataset that appear ambiguous or challenging to classify. We then ask 10 participants to determine whether the character in each image resembles the target character. These participants are pre-screened to ensure they are familiar with all the target characters. Once the responses are collected, we calculate an agreement score, with the majority opinion among the participants serving as the final label for each image.
\label{label img}

%\subsection{Conclusion}

\begin{figure*}[t]
    \centering
    \includegraphics[width=1\linewidth]{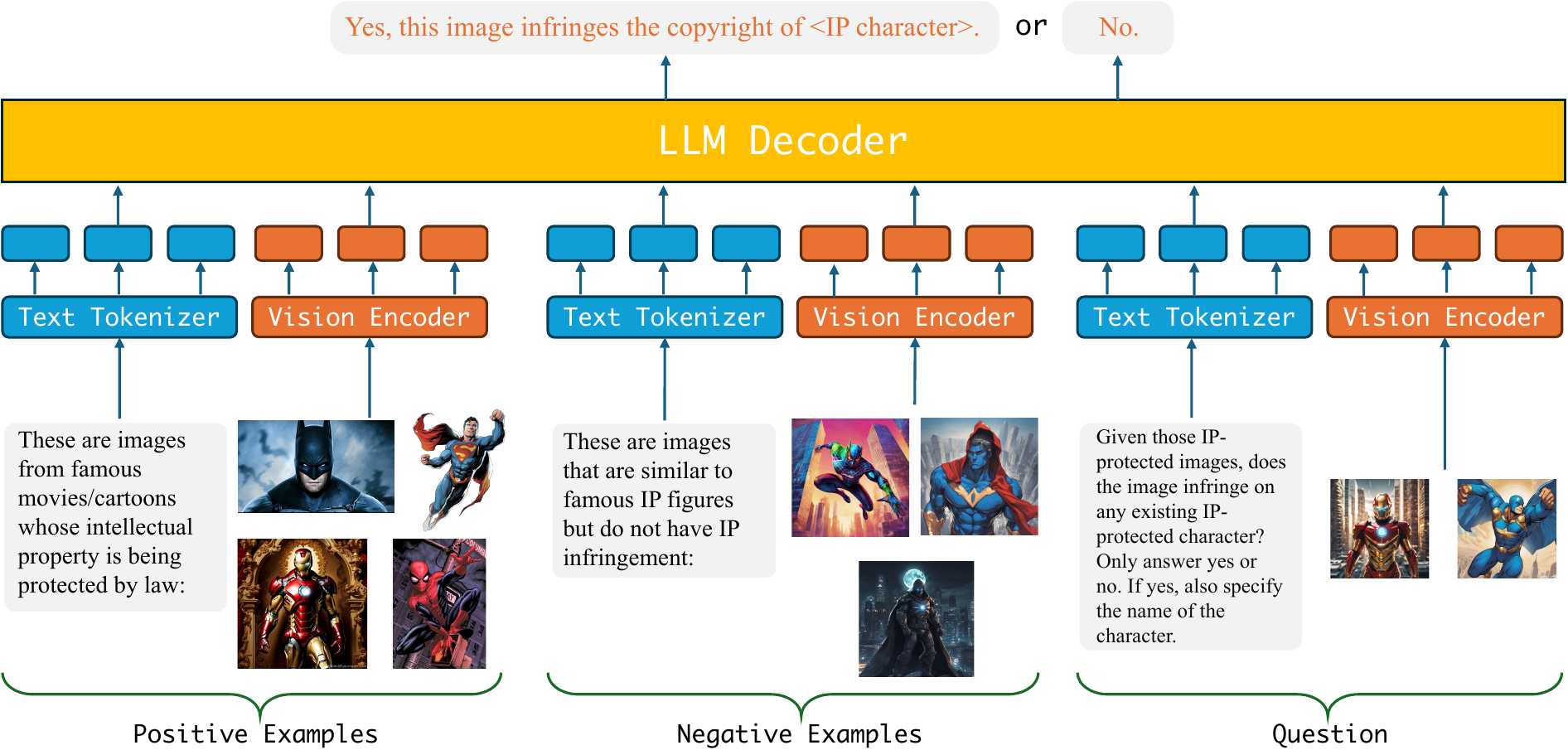}
    % \vskip -0.5cm
    \caption{Examples of detecting IP infringements using in-context learning}
    \label{fig:6}
\end{figure*}

\begin{table*}[ht]
    \setlength{\tabcolsep}{2pt}
    \centering
    \footnotesize
    \begin{tabular}{lccccccccccc}
    \toprule
    & \multicolumn{2}{c}{\textbf{SpiderMan}} & \multicolumn{2}{c}{\textbf{BatMan}} & \multicolumn{2}{c}{\textbf{IronMan}} & \multicolumn{2}{c}{\textbf{SuperMan}} & \multicolumn{2}{c}{\textbf{SuperMario}}\\
    \cmidrule(lr){2-3} \cmidrule(lr){4-5} \cmidrule(lr){6-7} \cmidrule(lr){8-9} \cmidrule(lr){10-11}
    \textbf{Models} & ICL & 0-shot & ICL & 0-shot & ICL & 0-shot & ICL & 0-shot & ICL & 0-shot\\
    \midrule
       Claude 3.5 & \textbf{0.56/1} & 0.42/1 & \textbf{0.57/1} & 0.51/1 & \textbf{0.44/1} & 0.43/1 & \textbf{0.56/1} & 0.55/1 & \textbf{0.75/1} & 0.64/1 \\
       GPT-4o & \textbf{0.57/1} & 0.43/1 & \textbf{0.39/1} & 0.37/1 & \textbf{0.49/1} & 0.44/1 & 0.64/0.99 & \textbf{0.65/0.99} & \textbf{0.74/1} & 0.71/0.91\\
       GPT-4o mini & \textbf{0.58/1} & 0.41/0.97 & \textbf{0.54/0.97} & 0.4/0.7 & \textbf{0.52/0.93} & 0.45/0.91 & \textbf{0.67/1} & 0.567/1 & \textbf{0.82/0.93} & 0.75/0.88\\
       VILA-2.7b & \textbf{0.55/0.78} & 0.41/1 & \textbf{0.63/0.79} & 0.6/0.76 & \textbf{0.43/0.95} & 0.38/0.93 & 0.53/1 & \textbf{0.55/0.99} & 0.68/0.95 & \textbf{0.7/0.86}\\
       Qwen-VL-7b & \textbf{0.48/0.8}& 0.4/0.75 & \textbf{0.55/1} & 0.51/0.95 & \textbf{0.45/1} & 0.43/1 & \textbf{0.52/1} & 0.5/1 & \textbf{0.78/0.98} & 0.7/0.95\\
       DeepSeek-VL2-1b & \textbf{0.5/0.9} & 0.45/0.85 & \textbf{0.52/1} & 0.48/1 & 0.47/0.96 & 0.48/0.95 & \textbf{0.6/1} & 0.56/1 & \textbf{0.7/1} & 0.63/0.96 \\
       Intern-VL2-2b & \textbf{0.5/0.9} & 0.45/0.8 & \textbf{0.58/0.9} & 0.48/0.92 & \textbf{0.42/1} & 0.4/1 & \textbf{0.6/1} & 0.58/0.98 & \textbf{0.72/0.95} & 0.71/0.98\\

    \bottomrule
    \end{tabular}
    \caption{Precision/Recall value from different models incorporated with in-context learning and with zero-shot VQA. Each model is evaluated from different IP figures of our benchmark dataset.}
    \label{table2}
\end{table*}

\section{VLM IP Infringement Detection Experiments}
%In this section, we evaluate five vision-language models (VLMs) for their intellectual property infringement detection capabilities: GPT-4o \citep{gpt4v}, GPT-4o mini \citep{gpt4v}, Claude 3.5 \citep{Claude3.5}, LLaVA-2.7b \citep{liu2024visual}, and Qwen-VL \citep{bai2023qwen}. Our assessment employs two approaches—one utilizing in-context learning and the other without—applied to our benchmark dataset. 

\subsection{Experimental Settings}
\textbf{Dataset.} In our experiments, we use our benchmark dataset as discussed in Sec. \ref{Sec 3}, which contains famous IP characters: Iron-Man, Batman, Spider-Man, Superman, and Super Mario, with each class contains challenging hard negative samples.

\noindent
\textbf{Large Vision Language Models.} We evaluate seven large vision-language models (LVLMs) for their intellectual property infringement detection capabilities: GPT-4o \citep{gpt4v}, GPT-4o mini \citep{gpt4v}, Claude 3.5 \citep{Claude3.5}, VILA-2.7b \citep{lin2023vila}, Qwen-VL-7b \citep{bai2023qwen}, DeepSeek-VL2-1b \cite{wu2024deepseek} and Intern-VL2-2b \cite{chen2024internvl}. Developed by OpenAI, GPT-4o and GPT-4o mini are multilingual, multimodal generative pre-trained transformers capable of processing and generating text, images, and audio. Anthropic's Claude 3.5 Sonnet is an AI model that boasts impressive benchmark scores. VILA-2.7b is a vision-language model designed to integrate visual and textual information for comprehensive understanding and generation tasks.  InternVL2 is a pioneering open-source alternative to GPT-4o. Qwen-VL and DeepSeek-VL2 are state-of-the-art multimodal models that rivals Claude Sonnet and GPT-4o, with open weights. We compare and analyze the infringement detection capability of these models on our benchmark dataset using in-context learning and zero-shot VQA.

\noindent
\textbf{Evaluation Metrics.}
To evaluate the IP infringement detection capabilities of LVLMs on our benchmark dataset, we employed a diverse set of evaluation metrics that capture different aspects of their performance. These metrics provide a well-rounded assessment of each model’s strengths and weaknesses in identifying IP infringement cases.  

We assess the LVLM detection performance using the following key metrics:  

\noindent
\textbf{Precision}: Measures the proportion percentage of correctly identified infringement cases among all detected cases.

\noindent
\textbf{Recall}: Evaluates the proportion of actual infringement cases that the model successfully identifies.

By analyzing these metrics together, we gain deeper insights to help us understand the models’ effectiveness in accurately detecting copyright violations.

\subsection{Evaluate with In-context Learning}
To effectively utilize in-context learning for IP infringement detection, we begin by providing the vision-language model with a set of labeled examples, including both positive and negative instances of intellectual property (IP) content. These examples are accompanied by text prompts that clearly indicate whether they are positive samples (legally protected IP) or negative samples (content free from copyright concerns). This approach helps the model learn to distinguish between protected and non-protected contents within the given context. Subsequently, we give the LVLMs an image sample, and query whether the image infringes any existing IP-character. The output from the vision-language model serves as the final determination for identifying potential infringement in the given image, as shown in Fig. \ref{fig:6}. This approach leverages the model's ability to analyze and interpret visual content in conjunction with textual information, enabling it to assess whether the image contains elements that may violate intellectual property rights. By evaluating the image through this model, it is possible to detect unauthorized use of copyrighted material or other forms of infringement. 

\subsection{Evaluate with Zero-shot VQA}
We also investigate the IP infringement detection with zero-shot VQA. This is achieved by directly asking vision-language models whether the given image samples has copyright issues. 

\subsection{Experimental Results and Analysis}
In Table \ref{table2}, we evaluate the performance of LVLMs using in-context learning (ICL) and compare it with their performance with zero-shot learning across various intellectual property (IP) metrics from our benchmark dataset. The results demonstrate that IP infringement detection with in-context learning significantly outperforms detection with zero-shot learning, highlighting the effectiveness of integrating ICL with LVLMs for this task. Among the evaluated models, GPT-4o mini achieves the highest performance in IP infringement detection. Additionally, the results reveal that all models maintain a high recall score but exhibit relatively low precision. This suggests that LVLMs are more sensitive to positive models than negative models, and tend to classify image samples as positive cases more frequently, potentially leading to the misidentification of numerous negative samples as instances of IP infringement.

\begin{table*}[t!]
    \centering
    \footnotesize
    \renewcommand{\arraystretch}{1.5} % Adjust row height
    \begin{tabular}{lcp{6cm}} % Set fixed width for the last column
    \toprule
    FP image samples & Negative class & Reason for identifying it as a positive sample\\
    \midrule
    \raisebox{-0.8\totalheight}{\includegraphics[width=0.12\linewidth]{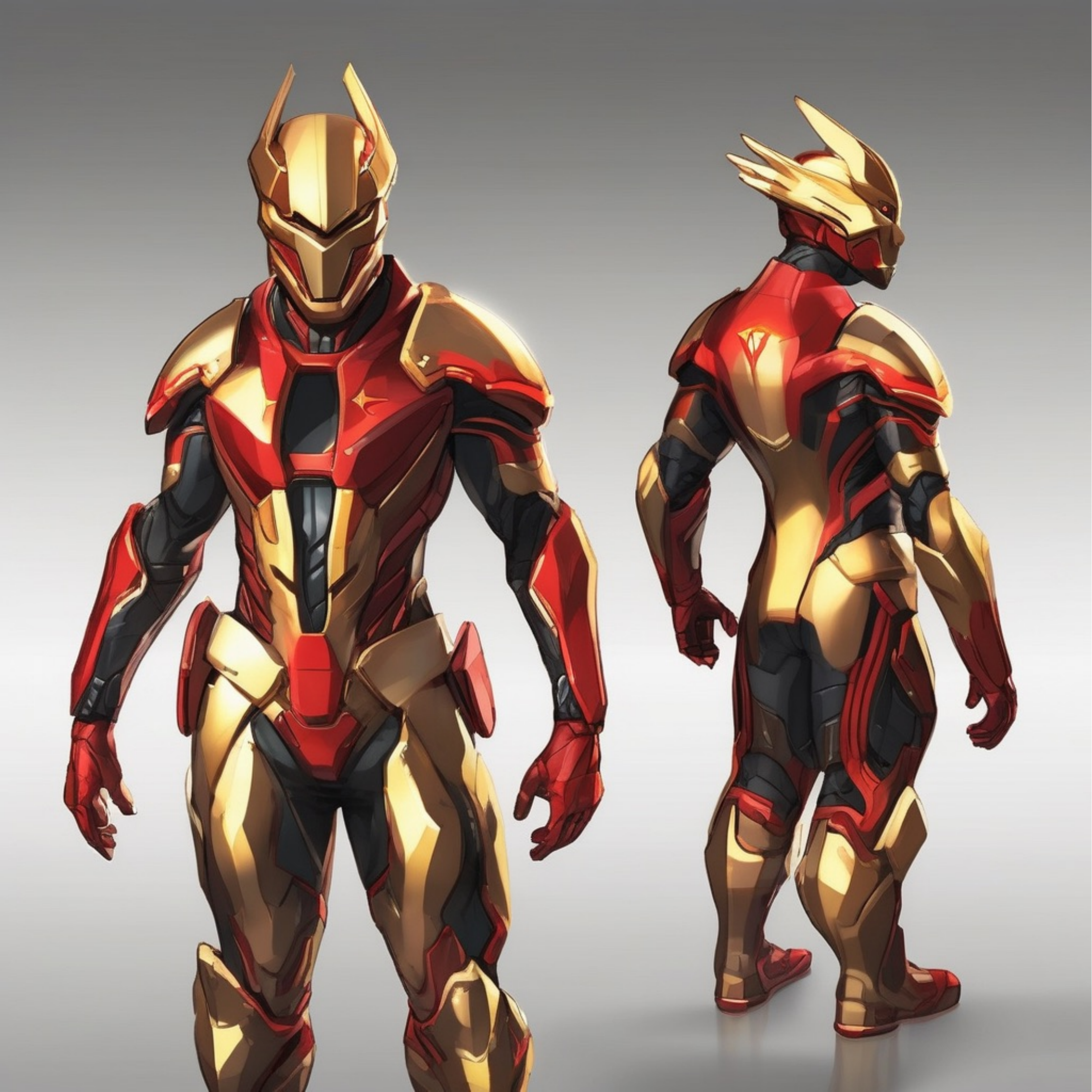}} & 
    IronMan & The armor in the image is similar to the golden armor worn by Ironman. The color and shape of the armor closely resemble the golden-red suit seen in various appearances of the character. \\
    \midrule
    \raisebox{-0.8\totalheight}{\includegraphics[width=0.12\linewidth]{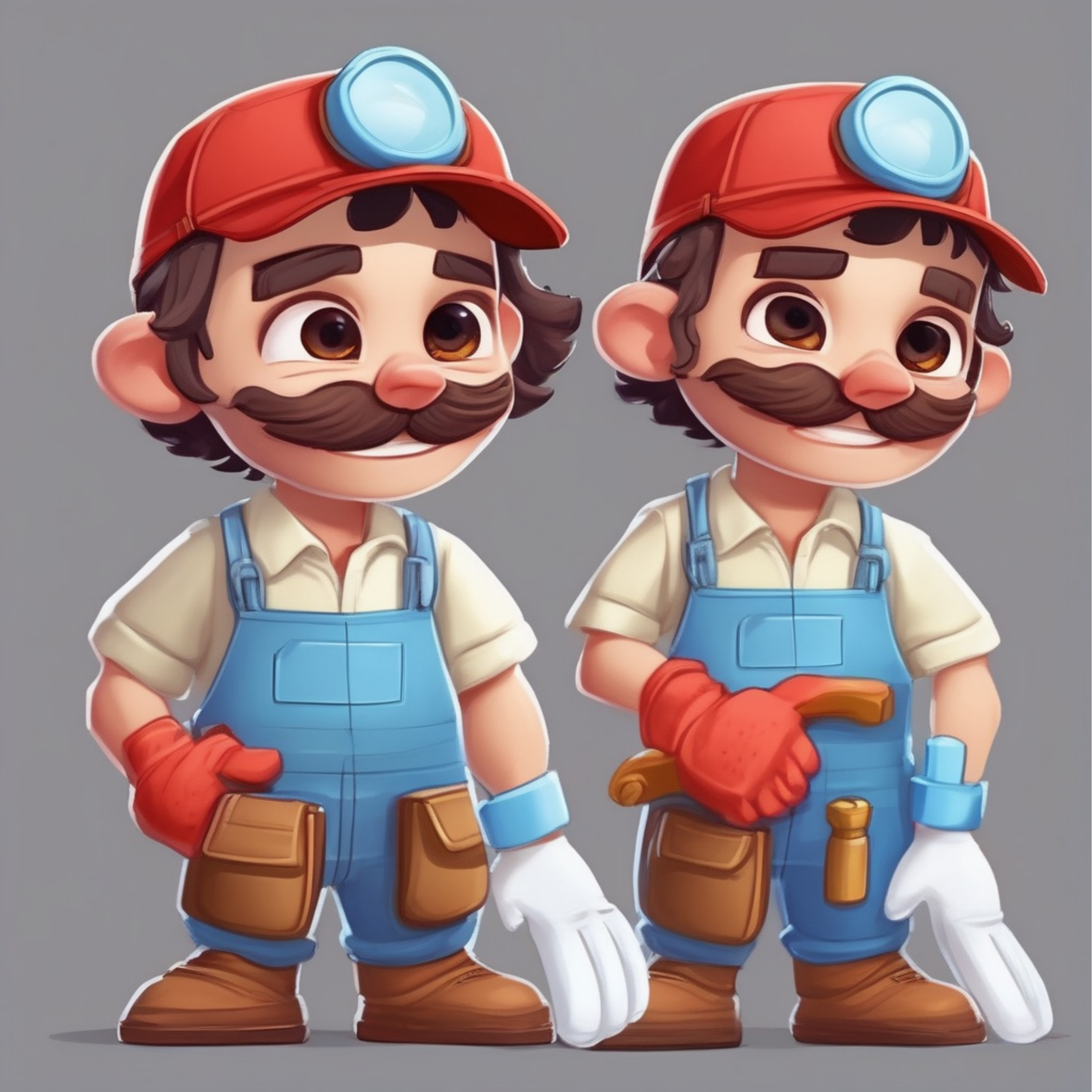}} &
    SuperMario & It showcases a character with a similar hat, mustache, overalls, and gloves, containing most of the features of the character SuperMario. \\
    \midrule
    \raisebox{-0.8\totalheight}{\includegraphics[width=0.12\linewidth]{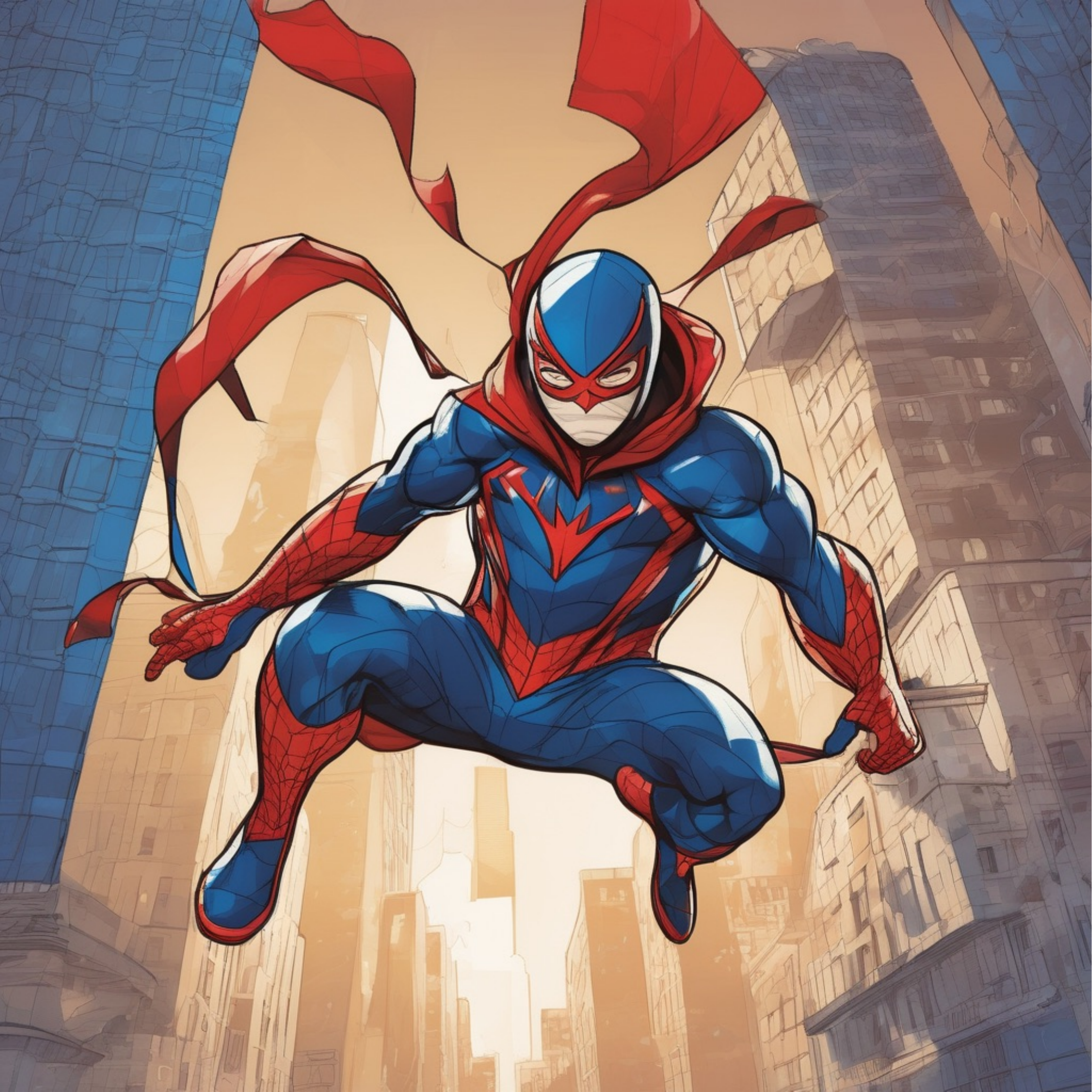}} &
    SpiderMan & The character is positioned in a dynamic web-slinging pose, which is strongly associated with Spider-Man's signature acrobatic movements. \\
    \midrule
    \raisebox{-0.8\totalheight}{\includegraphics[width=0.12\linewidth]{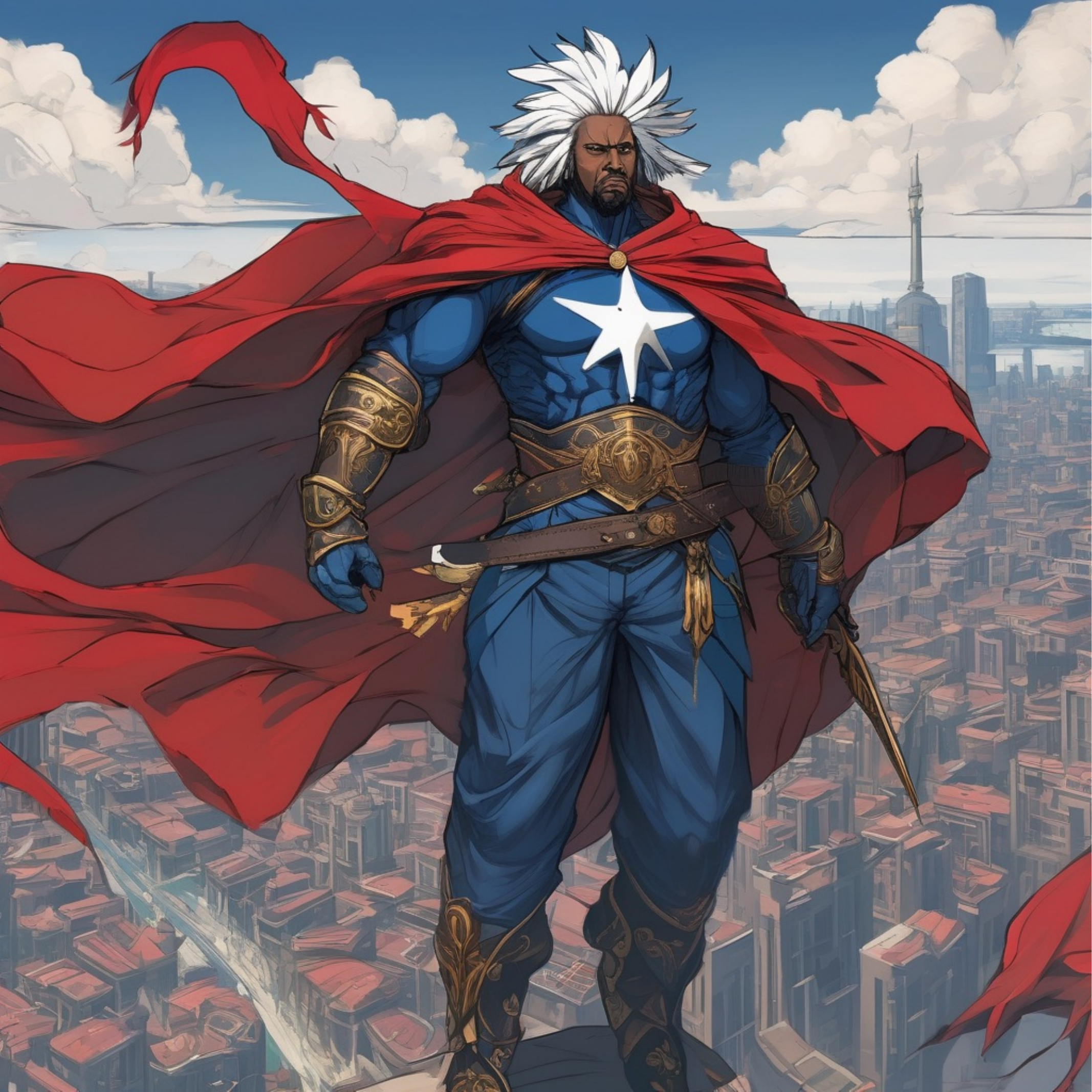}} & 
    SuperMan & The image contains features like a strong muscular build, a cape, and a prominent costume with a distinct emblem, containing most of the features of the character SuperMan. \\
    \midrule
    \raisebox{-0.8\totalheight}{\includegraphics[width=0.12\linewidth]{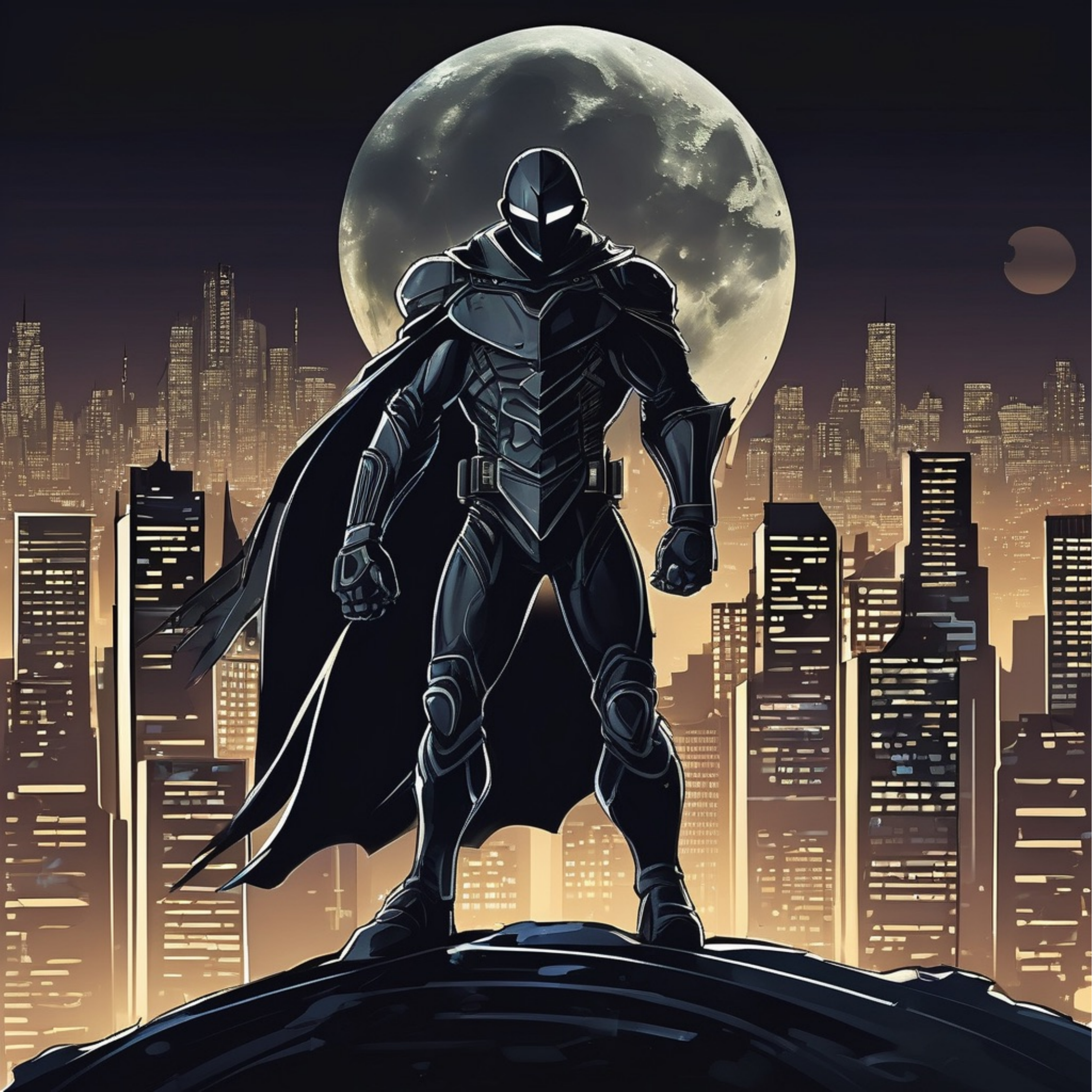}} & 
    BatMan & The image below contains features of Batman, notably his iconic cape, mask, and the dark color palette that aligns with the character's aesthetic. \\
    \bottomrule
    \end{tabular}
    \caption{False positive image samples and their negative classes detected by GPT-4o mini. The last column shows the reason why GPT-4o mini identifies them as positive samples.}
    \label{table3}
\end{table*}

\section{Failure Case Analysis}
In this section, we analyze the failed detection cases produced by LVLMs. The results in Table \ref{table2} indicate that all LVLMs tend to exhibit high recall values while maintaining relatively low precision. This suggests that the primary source of misjudgment in LVLMs is false positive samples. Consequently, we delve deeper into false negative cases to understand their underlying causes and explore potential solutions.

We specifically examine false positive samples from the GPT-4o mini model combined with in-context learning and analyze the reasoning behind its classification of images as positive cases, as shown in Table \ref{table3}. The results reveal that the GPT-4o mini model primarily focuses on specific features, such as the "golden armor" worn by Iron Man, the "mustache" of Super Mario, and the "muscular build with a red cape" associated with Superman. While these negative samples exhibit prominent characteristics found in well-known intellectual property (IP) figures, they do not necessarily constitute IP infringement, as they lack a strong overall resemblance to the original characters. To address the aforementioned issues, one possible solution is to fine-tune the pretrained LVLMs on hard negative samples—instances where the model is prone to errors—using contrastive learning, such as CLIP’s loss \citep{pmlr-v139-radford21a}.

\section{Conclusion}
In this paper, we present a novel benchmark dataset specifically designed to evaluate the copyright detection capabilities of Large Vision-Language Models (LVLMs). Our dataset incorporates both positive and negative samples, carefully engineered and manipulated through input text prompts to rigorously assess model performance.
Furthermore, we conduct a comprehensive analysis of state-of-the-art LVLMs, including GPT-4o, Claude 3.5, Vila 2.7b, Qwen-VL-7b, DeepSeek-VL2-1b, and Intern-VL2-2b across diverse experimental settings. Our evaluation of six cutting-edge LVLMs reveals significant shortcomings in identifying and detecting intellectual property (IP) infringement, particularly when faced with challenging negative samples.
To further investigate these limitations, we analyze failure cases—specifically, false positives—by probing the reasoning behind LVLM misclassifications. Our findings suggest that while LVLMs focus on specific features of potentially infringing images, they lack the holistic judgment necessary for accurate copyright assessment.
These insights highlight a pressing need for dedicated benchmarks to guide the development and validation of copyright-specific LVLMs, ensuring their effectiveness in real-world applications.

\section{Limitations}
\textbf{Challenges in Dataset Annotations.} While this paper introduces a dataset containing IP infringement samples to facilitate the evaluation of LVLMs, it does not fully account for the legal thresholds of copyright infringement, which vary across jurisdictions and involve complex interpretations beyond mere visual similarity. For instance, U.S. copyright law considers factors such as substantial similarity and fair use when determining infringement \cite{joyce2016copyright, howell1942copyright, latman1980copyright}—an intricate legal assessment that goes beyond the scope of our dataset labeling process. Instead, we rely on common sense and consensus-based annotation, which, while practical, may introduce biases due to the absence of comprehensive legal and ethical considerations. Consequently, our dataset may not perfectly align with real-world copyright enforcement standards.

%\newpage
% Bibliography entries for the entire Anthology, followed by custom entries
%\bibliography{anthology,custom}
% Custom bibliography entries only
\bibliography{main}

\newpage
\appendix

\section{The Effectiveness of Non-overlappped Negative Prompt (NNP)}
In this section, we compare the performance of plain negative prompts and non-overlapped negative prompts. To evaluate text-to-image alignment, we utilize the CLIP score \cite{rando2022red}. Additionally, we assess the IP infringement rate through human evaluation, where a panel of five inspectors determines whether the generated images resemble the target IP characters. The results are presented in Table \ref{table4}.

\begin{table}[h]
\centering
\footnotesize
\caption{The CLIP Score of the undefended model, with plain negative prompt (PNP) and with non-overlapped negative prompt (NNP). The model used here is Stable Diffusion XL.}
\begin{tabular}{lcccc}
\toprule
Character  & Undefended & PNP &  NNP\\ \midrule
Spider-Man      & 34.17      & 30.14   & 32.73 \\
Iron Man        & 27.93      & 26.33   & 27.86 \\
SuperMario      & 35.49      & 32.27   & 33.41 \\
Batman          & 28.53      & 29.01   & 30.73 \\
Superman        & 32.22      & 30.80   & 32.01 \\ \bottomrule
\end{tabular}
\label{table4}
\end{table}

\begin{table}[h]
\centering
\footnotesize
\caption{The IP infringement rate of the undefended model, with plain negative prompt (PNP) and with non-overlapped negative prompt (NNP). The model used here is Stable Diffusion XL.}
\begin{tabular}{lcccc}
\toprule
Character  & Undefended & PNP &  NNP\\ \midrule
Spider-Man      & 76.6      & 5.8   & 5.0 \\
Iron Man        & 48.6      & 0.0   & 0.0 \\
SuperMario      & 50        & 0.0   & 0.0 \\
Batman          & 50.8      & 1.6   & 0.0 \\
Superman        & 93.8      & 6.4   & 4.0 \\ \bottomrule
\end{tabular}
\label{table5}
\end{table}

\begin{table}[t]
\centering
\footnotesize
\caption{SAM+VLM with zero-shot VQA.}
\begin{tabular}{lcc}
\toprule
Character  & Precision & Recall \\ \midrule
Spider-Man & 0.4      & 0.9   \\
Iron Man   & 0.4      & 0.92   \\
SuperMario & 0.65     & 0.9   \\
Batman     & 0.6      & 1   \\
Superman   & 0.55     & 0.8 \\ \bottomrule
\end{tabular}
\label{table7}
\end{table}

\begin{table*}[t]
    \centering
    \scriptsize
    \renewcommand{\arraystretch}{1.5} % Adjust row height
    \begin{tabular}{lcccp{7cm}} % Set fixed width for the last column
    \toprule
    Image samples & Class & Generative Models & Negative Prompts/Weights &Positive Prompts\\
    \midrule
    \raisebox{-0.8\totalheight}{\includegraphics[width=0.13\linewidth]{figures/Descriptive_prompt/ironman.pdf}} & 
    IronMan & Stable Diffusion XL & N/A & A futuristic suit-clad hero, featuring a sleek red and gold armor with a glowing chest reactor. This character flies with jet-powered boots, has advanced technology built into the suit, and showcases a heroic stance against a modern city backdrop. The design merges cutting-edge science with undeniable charisma.
    \\
    \midrule
    \raisebox{-0.8\totalheight}{\includegraphics[width=0.13\linewidth]{figures/Descriptive_prompt/batman.pdf}} &
    BatMan & Ideogram AI & N/A & Design a masked vigilante at night, wearing a dark, armored suit with a cape. He stands atop a gothic cityscape, poised dramatically. His gear includes advanced gadgets on his utility belt, and his eyes emit a fierce determination. The mood is gritty and mysterious, under a shadowy, cloud-filled sky.
    \\
    \midrule
    \raisebox{-0.8\totalheight}{\includegraphics[width=0.13\linewidth]{figures/Descriptive_prompt/superman.pdf}} &
    SuperMan & Stable Diffusion XL & N/A & Create an image of a muscular male superhero with a blue suit, red cape, and emblem on his chest. His hair is neatly combed back, and he's flying above a bustling city. His eyes glow as he looks down, ready to swoop in and save the day.
    \\
    \midrule
    \raisebox{-0.8\totalheight}{\includegraphics[width=0.13\linewidth]{figures/Descriptive_prompt/spiderman.pdf}} &
    SpiderMan & Stable Diffusion XL & N/A & Design an image of a youthful superhero in a sleek, red and blue suit with arachnid-inspired motifs. The character exhibits agility and is posed dynamically, perhaps swinging between skyscrapers with web-like strands emanating from his wrists, set against an urban backdrop.\\
    \midrule
    \raisebox{-0.8\totalheight}{\includegraphics[width=0.13\linewidth]{figures/Descriptive_prompt/supermario.pdf}} &
    SuperMario & Ideogram AI & N/A & Visualize a stout, cheerful man in blue overalls, a red cap, and shirt, sporting a thick mustache. He exudes an adventurous spirit, often depicted with a gleeful smile, jumping energetically. His background is usually brimming with vibrant, whimsical landscapes reminiscent of classic video game worlds. \\
    \midrule
    \raisebox{-0.8\totalheight}{\includegraphics[width=0.13\linewidth]{figures/negative_prompt/batman.pdf}} & BatMan
    & DALL-E & PNP/-1.0 & Design a masked vigilante at night, wearing a dark, armored suit with a cape. He stands atop a gothic cityscape, poised dramatically. His gear includes advanced gadgets on his utility belt, and his eyes emit a fierce determination. The mood is gritty and mysterious, under a shadowy, cloud-filled sky.\\
    \midrule
    \raisebox{-0.8\totalheight}{\includegraphics[width=0.13\linewidth]{figures/Perp_Neg/batman.pdf}} & BatMan
    & DALL-E & NNP/-1.0 & Design a masked vigilante at night, wearing a dark, armored suit with a cape. He stands atop a gothic cityscape, poised dramatically. His gear includes advanced gadgets on his utility belt, and his eyes emit a fierce determination. The mood is gritty and mysterious, under a shadowy, cloud-filled sky.\\
    \midrule
    \raisebox{-0.8\totalheight}{\includegraphics[width=0.13\linewidth]{figures/negative_prompt/ironman.pdf}} & IronMan
    & DALL-E & PNP/-1.0 & Design an armored superhero with a sleek, red and gold exosuit, equipped with repulsor technology and flight capabilities. The suit features an arc reactor on the chest and is sophisticated, embodying futuristic combat and defense technology. This character is charismatic and has a genius-level intellect, embodying innovation and leadership.\\
    \midrule
    \raisebox{-0.8\totalheight}{\includegraphics[width=0.13\linewidth]{figures/Perp_Neg/ironman.pdf}} & IronMan
    & DALL-E & NNP/-1.0 & Design an armored superhero with a sleek, red and gold exosuit, equipped with repulsor technology and flight capabilities. The suit features an arc reactor on the chest and is sophisticated, embodying futuristic combat and defense technology. This character is charismatic and has a genius-level intellect, embodying innovation and leadership.\\
    \bottomrule
    \end{tabular}
    \caption{Image samples generated with different generative models and different prompts.}
    \label{table6}
\end{table*}

\section{Dataset Images and Their Corresponding Prompts}
\label{sec:appendix}
In this section, we will display dataset images and their corresponding prompts. 
\subsection{Descriptive Prompt Generation}
    We generate the descriptive prompt by employing a large language model. And we use GPT-4o \cite{gpt4v} here as its exceptional text generation capabilities. We construct descript prompt with the following guidance to GPT-4o:
    \begin{itemize}
        \item Creating a prompt that describes a character similar to <Target
    Character>. This prompt should enable text-to-image AI models
    to generate images without directly mentioning the name of the
    <Target Character>.
    \end{itemize}

\subsection{Visualization}
In this section, we will display image samples with their corresponding classes, input prompts, and generative models which produced them, as shown in Table \ref{table4}.

\section{Prompt for Failure Case Analysis}
The following prompts are used to identify and analyze failed false positive cases, as discussed in Section 5:

\begin{itemize}
        \item Could you provide a plausible reason for identifying this image as a violation of the intellectual property rights of <Target IP Character>?
\end{itemize}

\section{Possible Solutions for Improving IP Infringement Detection}
Since IP figures often include background elements that may interfere with LVLMs in detecting potential IP infringements, one possible approach to enhancing detection is leveraging the Segment Anything Model (SAM)—a state-of-the-art image segmentation model. Specifically, we apply SAM with text prompts corresponding to the IP figures to generate masks that isolate these figures within the images. We then follow the procedures outlined in Section 4 using images that contain only the masked IP figures.  

Experimental results in Table \ref{table7} indicate a slight improvement in precision and recall when combining SAM with LVLMs. However, the enhancement is marginal, and the approach lacks significant innovation and novelty. These findings suggest that achieving substantial advancements in IP infringement detection remains a challenging problem in this domain.

\end{document}